\pdfoutput=1
\documentclass[11pt]{article}

\usepackage{acl}

\usepackage{times}
\usepackage{latexsym}

\usepackage[T1]{fontenc}
\usepackage[utf8]{inputenc}

\usepackage{microtype}

\usepackage{inconsolata}

\usepackage{booktabs}
\usepackage{graphicx}
\usepackage{makecell}
\newcommand{\eat}[1]{}
\usepackage{subcaption}
\usepackage{amsmath}
\usepackage{enumitem}
\usepackage{tabularx}

\newcommand{\edit}[1]{\textcolor{blue}{#1}}
\newcommand{\todo}[1]{\textcolor{red}{[\textsc{TODO: }#1 ]}}
\newcommand{\dataset}[0]{DS Critique Bank}
\newcommand{\model}[0]{Digital Socrates}
\newcommand{\modelabb}[0]{DS}
\newcommand{\dataseturl}[0]{\url{https://allenai.org/data/digital-socrates}}
\title{Digital Socrates: Evaluating LLMs through Explanation Critiques}
\author{
Yuling Gu, Oyvind Tafjord, Peter Clark \\
Allen Institute for AI, Seattle, WA \\
\texttt{\{yulingg,oyvindt,peterc\}@allenai.org} 
}

\begin{document}
\maketitle
\pagestyle{plain} %ADD FOR page numbers - REMOVE FOR ACL%%
\begin{abstract}
While LLMs can provide reasoned
explanations along with their answers,
the nature and quality of those explanations are still poorly understood.
In response, our goal is to define a detailed way of characterizing the
explanation capabilities of modern models and to
create a nuanced, interpretable explanation evaluation tool that can
generate such characterizations automatically,
without relying on expensive API calls or human annotations.
Our approach is to (a) define the new task of \textit{explanation
critiquing} - identifying and categorizing any main flaw in an explanation
and providing suggestions to address the flaw, (b) create a sizeable,
human-verified dataset for this task, and (c) train an open-source,
automatic critique model (called Digital Socrates) using this data. Through
quantitative and qualitative analysis, we demonstrate how
Digital Socrates is useful for revealing insights about student models by
examining their reasoning chains, and how it can provide high-quality,
nuanced, automatic evaluation of those model explanations for the first
time. Digital Socrates thus fills an important gap in evaluation
tools for understanding and improving the explanation
behavior of models.
\eat{
[[EARLY PLACEHOLDER ABSTRACT, WILL BE REPLACED]]\\
Current NLP models based on large language models (LLMs), such as GPT-4 and Llama-2, have the ability to provide explanations along with their answers, for instance through chain-of-thought prompting. We explore how these explanations can provide a deeper window into capabilities of the model. Taking inspiration from Socratic principles of effective tutoring, we design a critique format which gives feedback on a model's explanation. The feedback identifies and categorizes any main flaw, and provides both general and specific suggestions to fix it. We generate such critiques both by prompting a very capable model, like GPT-4, and by fine-tuning a smaller Llama2 models using gold critiques. We show that such critiques can show interesting contrasts between models (like GPT-4 vs Llama2-70B) beyond pure accuracy evaluations. We also show that the smaller critique model is more effective on certain types of critique dimensions (like sloppy reasoning) than others (like long-tail knowledge). Finally, the critiques can be used as effective feedback to a range of "student" models in improving their output, even when the student model is larger than the critique model.}

\end{abstract}

% The unexamined life is not worth living—Socrates
% So we need to examine our explanations!

% Figure 1
%https://docs.google.com/presentation/d/1YELroK3iaCvfixrhcxR9ld496r8PSagwBLbp9f76cc0/edit#slide=id.p

\begin{figure}[t!]
\centering
      \includegraphics[width=\columnwidth]{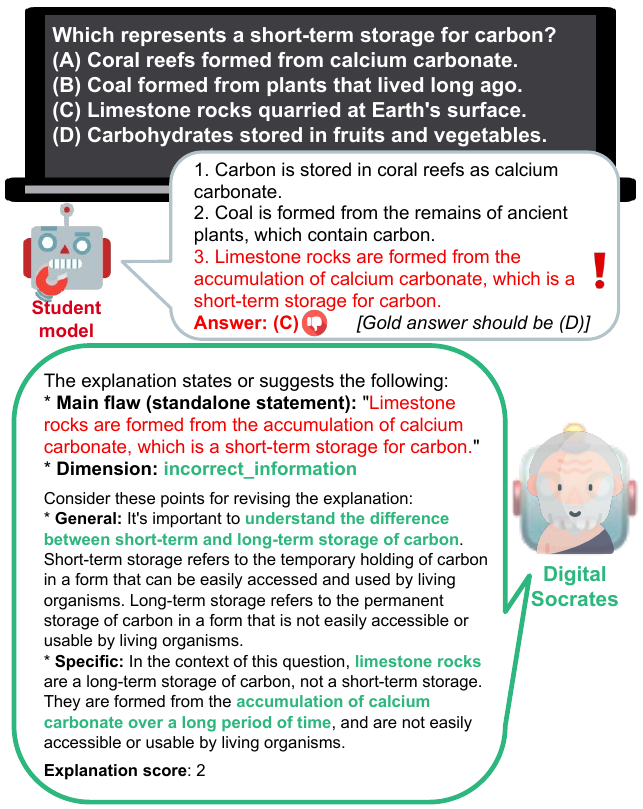}
\caption{Given a multiple-choice question (together with the answer options and correct answer), as well as a model-generated reasoning chain and answer, our system \model{} gives a critique of the model-generated explanation. In its critiques, \model{} provides localized feedback on where and why reasoning chains are flawed (focusing on the main flaw, if any), accompanied by general and fine-grained suggestions to address the identified flaw, providing nuance and interpretability to the critiques.}
\label{fig:ds_summary}
\vspace{-4mm}
\end{figure}

\section{Introduction}

% introduction and motivation
Large language models (LLMs) have demonstrated promising end-task performance on a range of tasks \citep{openai2023gpt4, bubeck2023sparks}. These models, given their text-generation abilities, can also be prompted \citep{Wei2022CoTNeurips, Kojima2022ZeroShotNeurips} or trained \citep{rajani-2019-explain, li-etal-2023-symbolic, tafjord2021generalpurpose, gu-etal-2022-dream} to externalize their reasoning\footnote{A model-generated chain of reasoning can be a free-form explanation or a series of intermediate steps. We use this as a window into the model’s ability to create systematic arguments, but without making any claims about how LLMs
reason internally \cite{lm-humans-1,lm-humans-2}.}
as a window into their reasoning capabilities. Despite promising end-task performance, examining LLMs' reasoning chains
%\footnote{In the context of this work, model-generated explanations are viewed as a way of materializing the intermediate reasoning chains. For the rest of the paper, we use the terms ``intermediate reasoning chains" and ``explanations" interchangeably.} 
reveals gaps in the correctness of their factual knowledge and the coherence of their reasoning \citep{radhakrishnan2023question, lanham2023measuring, NEURIPS2022_c4025018, cohen2023lm}. Such efforts delving deeper into the quality of model-generated intermediate reasoning chains enable us to advance our understanding of LLMs' strengths and weaknesses in different tasks beyond measuring their performance on standard benchmarks.

In existing NLP works, however, judging the quality of LLMs' intermediate reasoning chains is met with several challenges. Early practices for determining the quality of such intermediate generations %when performing question-answering (QA) tasks 
include (1) reporting end-task accuracy and (2) sampling a subset to perform human annotations on e.g., \citet{liu-etal-2022-rainier, gu-etal-2022-dream, li-etal-2023-symbolic}. Using end-task performance as a proxy for the quality of intermediate reasoning 
%chains 
can be problematic as 
%recent works have demonstrated that 
LLMs' final answers can be unfaithful to the generated intermediate reasoning \citep{radhakrishnan2023question, lanham2023measuring}, whereas relying on human annotations %can be 
is labor-intensive and expensive. 

To tackle such challenges,
%in judging the quality of model-generated reasoning
several automatic measurements have been proposed. One category focuses on %metrics 
evaluating model-generated text based on a given reference text \textit{y} \citep{saunders2022selfcritiquing, kim2023prometheus, xu2023instructscore}.  Another 
%line of work 
proposes numerical metrics like ROSCOE \citep{golovneva2023roscoe} and ReCEval scores \citep{prasad2023receval} as a summary of %the 
reasoning quality. %Others build models that generate critiques and refinement suggestions in a free-form manner e.g., \citet{wang2023shepherd}.
Others build models that generate free-form critiques and refinement suggestions e.g., \citet{wang2023shepherd}. Our work builds upon these previous efforts, introducing a way of automatically evaluating reasoning chains that (1) focuses on the intrinsic quality of the reasoning chains, moving away from the reliance on comparing to any reference reasoning chain; (2) localizes where the reasoning went wrong and provides interpretable feedback on why that part of the reasoning chain should be revised; and
% , beyond providing a numeric rating on explanation quality; and 
(3) uses a semi-structured format useful for gaining both quantitative and qualitative insights about the reasoning chain quality.

To operationalize this, our approach and contributions are thus as follows: 
\begin{itemize}[noitemsep, topsep=0pt]
  \item We define the task of \textit{explanation critiquing}.
  \item We create \dataset{}, a sizeable, human-verified dataset 
  for the task, both to train critique models and to compare against explanation capabilities of future models.
  \item We train and release a high-performing, open-source critique model, \model{}, that does not rely on expensive API calls or human annotations.
  \item We demonstrate the usefulness of \model{} critiques.
\end{itemize}
By providing high-quality, nuanced automatic evaluation of explanations, \model{} fills an important gap in evaluation tools for the community. 
We make our dataset and model publicly available at \dataseturl{}.%\footnote{Anonymous release for reviewing stage: attached zip file.}
% Our dataset and model is available at []

\section{Related Work}
% Prompting or training these LLMs to externalize their reasoning as they answer a question has emerged as a promising solution not only to improve end-task performance but also, importantly, as a window into the reasoning capabilities of the models.

%\todo{Needs work}
% 
\textbf{LLM Evaluation.}
End-task performance on popular tasks is a standard way of evaluating models in NLP. For instance, Pythia \citep{biderman2023pythia} and  Llama-2 \citep{touvron2023llama} mostly focus on reporting accuracy on popular tasks like WSC \citep{Levesque2011TheWS}, ARC \citep{clark2018think} or benchmarks like MMLU \citep{hendrycks2021measuring}, and use that to guide model development. We evaluate explanations as an additional dimension.

\textbf{Explanations and Rationales.}
Free-form explanations and rationales have emerged as a promising approach to 
materialize models' reasoning chains 
%externalize models' reasoning chains as a window into their reasoning capabilities 
\citep{Wei2022CoTNeurips, Kojima2022ZeroShotNeurips, shwartz-etal-2020-unsupervised, gu-etal-2022-dream, chen-etal-2023-rev, wiegreffe-etal-2021-measuring}. %Previous works like \citet{chen-etal-2023-rev, wiegreffe-etal-2021-measuring} studied the relationship between rationales and labels.
Despite LLMs' impressive accuracy on a range of tasks \citep{openai2023gpt4, lms-few-shot-learners}, examining model-generated explanations surfaces issues such as the utility of such explanations to humans \citep{joshi-etal-2023-machine}, the correctness of models' factual knowledge, and the coherence of their reasoning \citep{radhakrishnan2023question, lanham2023measuring, NEURIPS2022_c4025018, cohen2023lm}. We propose the \textit{explanation critiquing} task as a systematic way to examine models' explanation quality.

\textbf{Model-generated Feedback/Critique.}
% Relying on human evaluation to examine models' reasoning capabilities through their externalized reasoning chain can be costly, encouraging a range of automatic measurements as proxies. 
%Prior works utilizing model-generated feedback, like \citet{qi2023art} and \citet{madaan2023selfrefine}, focused on using them to improve task performance. 
Prior works like \citet{qi2023art} and \citet{madaan2023selfrefine} focused on utilizing model-generated feedback to improve task performance. 
In terms of using automatic feedback for analyzing models, ROSCOE \citep{golovneva2023roscoe} provides numeric scores covering semantic alignment, semantic similarity, logical influence, and language coherence; ReCEval scores \citep{prasad2023receval} focus on the correctness and informativeness of reasoning chains. %Contributing to these existing works, 
%on automatic feedback for evaluating explanations, 
Continuing this line of work in a more general and flexible manner,
our \textit{explanation critiquing} task involves nuanced and interpretable critiques beyond numeric scores. Different from works like \citet{wang2023shepherd} which 
%tuned a model specifically to critique responses and 
suggest refinements in free-from text, our critique design is informed by principles of Socratic questioning \citep{Brunschwig:03, Paul2006-PAUTTG-6}. Further, extending efforts on critiquing model generations utilizing reference text by \citet{saunders2022selfcritiquing, kim2023prometheus, xu2023instructscore}, we move away from any reliance on reference texts. 
\eat{To substantiate this task, we introduce \dataset{}, a dataset of nuanced critiques. Further, leveraging on the potential of LLMs to provide feedback on sophisticated texts \citep{LLM-Research-Feedback-2023}, we also train a small, open-source, high-performing model (\model{}) on this task.
}

% ROSCOE paper \citep{golovneva2023roscoe}
% irrelevant = redundant
% repetition = all words occurred in the source text
% (focus on scoring, not explainable in terms of specific error, detailed information for available for fine grained analysis)

%focus on scoring, not explainable in terms of specific error, detailed information for available for fine grained analysis. 

% % critique is free-form
% 7B model specifically tuned to critique responses and suggest refinements \citep{wang2023shepherd}

% critiquing when given reference y
% critique summaries \citep{saunders2022selfcritiquing}

% \citep{kim2023prometheus} (similarity that we also generate feedback and score, try to train smaller bigger)
% (but we don’t have the reference answer setup, is more flexible, and focus on explanation) added capability - if detects wrong reasoning associated with correct answer

% \citep{xu2023instructscore} (finetune Llama 7B, focused on cases like MT where there is a reference answer y)

% Maybe LLMs will soon help us with reviews (or preemptively tell us what the reviews will be) \citep{LLM-Research-Feedback-2023}

%cite some distillation papers

\section{Explanation Critiquing: Task Design}
\label{sec:explanation_critiquing}
% introduce this new task
% task formulation
We formalize the \textit{explanation critiquing} task in this section. Then in the rest of the paper, we describe how we collect a dataset for this task, build a small and high-performing model for this task, and finally apply the model to reveal interesting insights about QA models in an automatic manner. 

Given a question $Q$, along with a model-generated explanation $E_{pred}$ and answer $A_{pred}$, the task involves giving a \textbf{critique of the model-generated explanation}, %which can be 
expressed as a 5-tuple:
\begin{multline*}
    critique(Q,E_{pred},A_{pred}) = \\ <f_{loc},f_{dim}, s_{gen}, s_{spec},E_{SC}>
\end{multline*}

\noindent
where $f_{loc}$ is a natural language (NL) statement that localizes and describes the most significant flaw (if any) in the explanation, $f_{dim}$ categorizes $f_{loc}$ into one of 8 categories of flaws described in Section \ref{subsec: task_design_identify_flaw} (or ``None''), $s_{gen}$ and $s_{spec}$ are general and specific NL suggestions for correcting
the flaw, and $E_{SC}$ is a numeric rating (from 0 to 5) indicating the explanation quality. See example critique in Figure~\ref{fig:ds_summary}.

%Given a question, along with a model-generated explanation and answer, the task involves giving a \textbf{critique of the model-generated explanation} that provides localized feedback on the most significant flaw (if any) in the explanation, suggestions to address the identified flaw, as well as a numeric rating indicating the explanation quality. See example critique in Figure~\ref{fig:ds_summary}.
% Example where model get wrong

\subsection{Revealing Problems in the Explanation}
\label{subsec: task_design_identify_flaw}
%Part1: identify and categorize any main flaw in the generated explanation
%The ideal critique of a given answer and explanation should be helpful, like one that would be given by a knowledgeable tutor. 
We design the first component, $f_{loc}$, of the critique to identify the most significant flaw (if any) in the explanation, pinpointing explicitly which part of the explanation is problematic. The flaw should be formulated as a standalone erroneous statement, that can be understood without reference to the full question or explanation.\footnote{More details on format in Appendix \ref{appendix-details-task-flaw-localize}.} 
%MOVED TO APPENDIX
%When possible, we encourage quote(s) in this statement to better localize the flaw, where in the quote(s), "..." is used to leave out parts not relevant to the flaw, and to keep just the erroneous parts. The quote(s) should be from the explanation, except in the case of pointing out a misunderstanding of the question and/or answer options, where the quote should try to identify the part of the question or answer that was misunderstood.

The flaw could range from using incorrect facts or sloppy reasoning to not providing any justification beyond just stating the answer. Informed by the systematic and disciplined method of Socratic questioning \citep{Brunschwig:03}, we target the critique to serve as a way of questioning the given explanation ``to get to the truth of things, to open up issues and problems, to uncover assumptions, to analyze concepts ... and to follow out logical implications of thought.'' \citep{Paul2006-PAUTTG-6} Specifically, our critique focuses on flaws along the 8 dimensions ($f_{dim}$) detailed in Table \ref{tab:8dims}. These dimensions have been chosen to cover the different types of Socratic questions enumerated in \citet{Paul2006-PAUTTG-6} to rigorously examine if the explanation demonstrates an understanding of the question, uses valid reasoning and evidence, considers appropriate viewpoints and perspective, and derives logical implications and consequences.
%MOVED TO APPENDIX
%\footnote{Following the spirit of the types of Socratic questions \citep{Paul2006-PAUTTG-6}, we focus on more significant flaws and ignore more trivial criticisms e.g., upper-case/lower-case issues, small typos, and spacing inconsistencies.}

\begin{table*} [t]
\centering
{\small
\begin{tabular}{cl} 
\hline
\textbf{Critique dimension}& \textbf{Description of flaw in the explanation} \\ \hline
misunderstanding & apparent misunderstanding of the original question or answer choices \\
lack justification & just stating the final answer without any proper or informative justification \\
incorrect information & stating incorrect fact(s) or knowledge \\
missing information &  missing a crucial fact, knowledge, or perspective that should be considered \\
incorrect reasoning & an incorrect leap in the reasoning \\
incomplete reasoning & the reasoning doesn't lead all the way to the answer \\
inconsistent answer & the answer doesn't match the conclusion of the explanation \\
irrelevant & using irrelevant or redundant fact(s), knowledge, or reasoning \\
\hline
\end{tabular}
}
\caption{Critique dimensions in our \textit{explanation critiquing} task. These dimensions cover the different types of Socratic questions \citep{Paul2006-PAUTTG-6} to guide rigorous examination of explanation quality. \label{tab:8dims} }
\vspace{-2mm}
\end{table*}

\subsection{Suggestions toward Truth and Accuracy}
\label{subsec: task_design_suggesttions}
The critique also contains general and specific suggestions to ensure that each flaw identified is justified with a direction for improvement in mind, rather than just being overly critical.
The general suggestion,
%in the critique, 
$s_{gen}$, is a statement that addresses a likely misconception or confusion underlying the flaw. This should 
%be a suggestion without 
not directly give out a particular answer to the question. Such information could then be used in future efforts to improve broad failure types in the explanation model.
The specific suggestion, $s_{spec}$, is a more targeted statement to consider when revising the explanation, as a guide towards the right reasoning chain (to the right answer) for this particular question.

\eat{
For each explanation, we couple revealing the most significant flaw (Section \ref{subsec: task_design_identify_flaw}) with general and specific suggestions to correct the flaw. The first part of the suggestion $s_{gen}$ is a general statement that addresses a likely misconception or confusion underlying the flaw. This should be a suggestion without directly giving out a particular answer to the question. The next part $s_{spec}$ is a more specific statement to consider when revising the explanation, as a guide towards the right reasoning chain (to the right answer) for this particular question.

Designing such suggestions as part of the \textit{explanation critiquing} task offers several advantages. First, it ensures that each flaw identified is justified with a direction for improvement in mind, rather than just being overly critical. In addition, the general suggestion could offer useful information for future efforts to improve the explanation model in terms of its failures for this type of error, whereas the more specific suggestion is targeted at correcting the model's reasoning for the specific question.
}

\subsection{Explanation Score}
%In addition to localizing flaws (Section \ref{subsec: task_design_identify_flaw}) and offering points to consider (Section \ref{subsec: task_design_suggesttions}), which are designed to make the critique interpretable and explainable,
The \textit{explanation critiquing} task also involves providing a quantitative metric on the explanation quality. The explanation score $E_{SC}$ is on a scale from 0 to 5 as summarized in Table \ref{tab:explanation_score_0-5}.

\eat{
\begin{table} [t]
\centering
{\small
\begin{tabular}{cl} 
\toprule
\textbf{\makecell[c]{Explanation\\score ($E_{SC}$)}} & \textbf{\makecell[l]{Description of\\ explanation quality}} \\ \midrule
0 & Very wrong \\ 
\midrule
1 & \makecell[l]{Has something useful  \\ relevant to the question} \\ 
\midrule
2 & \makecell[l]{Makes some valid points, but is\\ mostly incorrect} \\ 
\midrule
3 & \makecell[l]{Has many correct elements, \\ but with significant flaws} \\ 
\midrule
4 & Mostly correct, with a minor flaw \\ 
\midrule
5 & Completely correct \\
\bottomrule
\end{tabular}
}
\caption{Descriptions for different explanation scores $E_{SC}$ in our \textit{explanation critiquing} task. Part of the task involves rating the explanation quality on a scale of 0 to 5, where 0 indicates the explanation is very bad and 5 means that the explanation is completely correct. \label{tab:explanation_score_0-5} }
\vspace{-3mm}
\end{table}
}

\begin{table} [ht]
\centering
{\small
\begin{tabular}{cl} 
\toprule
\textbf{$E_{SC}$} & \textbf{Description of explanation quality} \\ \midrule
0 & Completely wrong \\ 
1 & Has something useful relevant to the question\\ 
2 & Makes valid points, but mostly incorrect \\ 
3 & Many correct elements, but with significant flaws\\ 
4 & Mostly correct, with minor flaw \\ 
5 & Completely correct \\
\bottomrule
\end{tabular}
}
\caption{Descriptions for different explanation scores $E_{SC}$ in the \textit{explanation critiquing} task, used to rate explanation quality on a scale from 0 to 5.
%, where 0 indicates the explanation is very bad %and 5 means that the explanation is completely %correct. 
\label{tab:explanation_score_0-5} }
\vspace{-1mm}
\end{table}

% Focusing on the most significant flaw* (critique is more concise and targeted, also helps evaluation - good/bad)
% Flaw formulated as standalone erroneous statement (with quotes)* + dimension + general and specific statements* as hint
% need for explainable

%\section{Experimental Setup}

\section{Dataset of Explanation Critiques}
We introduce \dataset{},\footnote{Digital Socrates' Critique Bank.} a critique dataset where each instance comprises a multiple-choice question (together with answer options and correct answer), a model-generated explanation and answer (using a variety of different models), a critique of the model-generated explanation, as well as (any) human annotations collected on that instance. \dataset{} focuses on questions requiring reasoning,
in particular science and commonsense reasoning (Section \ref{sec:dataset_tasks}). 
The explanations are from different models in popular explanation styles  (Section \ref{sec:dataset_models_explanation_styles}). The generation of critiques as well as the process for crowdsourced and expert annotations are described in Section \ref{sec:dataset_critiques}.

\subsection{Tasks}
\label{sec:dataset_tasks}
To obtain questions that require reasoning about basic science concepts, we sample from the AI2 Reasoning Challenge (ARC) dataset \citep{clark2018think}, which contains natural, grade-school level, multiple-choice science questions. 
% MOVED TO APPENDIX
%The ARC dataset has both a Challenge Set and an Easy Set, where the Challenge Set contains only questions that are regarded as more difficult (answered incorrectly by retrieval-based and word co-occurence algorithms). 
For commonsense reasoning questions, we sample from the RAINBOW \citep{Lourie_Le_Bras_Bhagavatula_Choi_2021} suite of tasks comprising 6 multiple-choice question-answering datasets -- $\alpha$NLI \citep{chandra2020a}, CosmosQA \citep{huang-etal-2019-cosmos}, HellaSwag \citep{zellers-etal-2019-hellaswag}, Physical IQa \citep{Bisk_Zellers_Le_bras_Gao_Choi_2020}, Social IQa \citep{sap-etal-2019-social} and WinoGrande \citep{Sakaguchi_Le_Bras_Bhagavatula_Choi_2020}. 
%Table \ref{tab:dataset_composition_simple} summarizes the main statistics of our \dataset{}, with more details in Table~\ref{tab:dataset_composition} in 
For evaluation purposes, we create a separate ``dev'' (evaluation) partition containing
additional samples from ARC and RAINBOW, plus samples from two additional 
datasets (i.e., OOD wrt. training), namely 
OpenBookQA (OBQA) \citep{mihaylov-etal-2018-suit} and CommonsenseQA (CSQA) \citep{talmor-etal-2019-commonsenseqa}. 
Table \ref{tab:dataset_composition_simple} summarizes the main statistics of  \dataset{}, with more details on sample sizes and mixes in Appendix~\ref{appendix-details-dataset}.
% More details are in Appendix~\ref{appendix-details-dataset} on how the partitions in \dataset{} cover questions from these QA datasets. 
% MOVED TO APPENDIX
% The DSCB-train-crowd-anno partition was purposefully skewed towards challenging questions, by sampling such that the Llama-7B-chat student model had low 30\% accuracy (vs the natural 60\% accuracy).\footnote{We favor sampling from Rainbow-Dev because the dev and test partitions of datasets like CosmosQA and Social IQa are designed to be of higher quality. CosmosQA sampled instances from workers who excelled in the generation task for their dev and test, while Social IQa's dev and test were validated by workers a second time.}

%Potential/Unseen: Science (OBQA), Commonsense (CSQA)

\begin{table} [t]
\centering 
{\small
\begin{tabular}{lccc} 
\toprule
\textbf{Partition} & 
\textbf{{\#}Qs} &
\textbf{{\#}Critiques} &
\textbf{{\#}Annos} \\ 
\midrule
 \makecell[l]{DSCB-train-silver} & 3240 & 3240 & 0 \\
 \makecell[l]{DSCB-train-crowd-anno} & 270 & 3240 & 4040 \\
\makecell[l]{DSCB-train-expert} & 31 & 198 & 198 \\
 \makecell[l]{DSCB-dev-crowd-anno} & 550 & 19800 & 2430 \\
\bottomrule
\end{tabular}
}
%\caption{Main statistics of \dataset{} partitions, showing number of unique
%questions, generated critiques and human annotations. \label{tab:dataset_composition_simple} }
\caption{%Statistics of \dataset{} partitions.
\dataset{} partitions provide breadth across different questions (DSCB-train-silver) as well as depth in comprehensively covering different student models and explanation styles (DSCB-train-crowd-anno). It contains both crowdsourced (DSCB-train-crowd-anno) and expert (DSCB-train-expert) annotations. The dev set is also released (DSCB-dev-crowd-anno). More details are in Appendix~\ref{appendix-details-dataset}.
\label{tab:dataset_composition_simple} }
\vspace{-3mm}
\end{table}

\eat{
ARC-Rainbow-270-train: 100 ARC-Challenge-Train + 50 ARC-Easy-Train + 20*6 Rainbow-Dev. Each with 4 student models * 3 prompts → 3240 Q+A+E+Crit instances
(probably chose Rainbow-Dev because COSMOS QA sampled from workers that excelled in the generation task for dev+test, Social IQA dev+test is validated by workers a second time)

ARC-Rainbow-3240-train: 810 ARC-Challenge-train + 810 ARC-Easy-Train + 270*6 Rainbow-Train. Each with ONE of 4 student models * 3 prompts → 3240 Q+A+E+Crit instances

ARC-Rainbow-OBQA-CSQA-550: 100 ARC-Challenge-Test + 100 ARC-Easy-Test + 50*6 Rainbow-Dev (no overlap with earlier usage) + 50 OBQA-Test + 50 CSQA-Dev. Each with 4 student models * 3 prompts, and GPT4-critique7 + GPT4-critique7\_nogold
}

\subsection{Explanations from Models}
\label{sec:dataset_models_explanation_styles}
\subsubsection{Explanation Styles}
We prompt models to generate an explanation 
%as a way of externalizing their reasoning chain 
along with their answers when performing QA tasks. We use three different prompts (see Appendix \ref{appendix-explanation_prompts}) which cover explanation styles in the spirit of Zero-shot Chain-of-Thought reasoning (CoT) \citep{Kojima2022ZeroShotNeurips}, Few-shot CoT \citep{Wei2022CoTNeurips}, and reasoning steps \citep{dalvi-etal-2021-explaining}.

\subsubsection{Explanation (``Student'') Models}
We use the following models to generate such explanations: GPT-4-0613 \citep{openai2023gpt4}, GPT-3.5-turbo-0613 \citep{lms-few-shot-learners}, Llama2-7B-chat, and Llama2-70B-chat \citep{touvron2023llama}. This covers strong proprietary models and open models of different strengths. For the rest of the paper, we will refer to these models whose outputs are to be critiqued as \textbf{student models}.

\subsection{Explanation Critiques}
\label{sec:dataset_critiques}
%The task of \textit{explanation critiquing} as formalized in Section \ref{sec:explanation_critiquing} can be laborious if authored from scratch.
%\footnote{Some of the authors tried writing such critiques from scratch and estmate that it takes 5-10minutes per question. } 
We elicit seed explanation critique data from GPT-4 \citep{openai2023gpt4}\footnote{Latest model at experimentation, GPT-4-0613.} and perform expert\footnote{Annotated by two of the authors.} and crowdsourced annotations on our training data (see Table \ref{tab:dataset_composition_simple}). The expert annotation involves (i) rating the GPT-4 generated critique (according to scales in Table \ref{tab:critique_score_0-3}) and (ii) revising bad critiques to make them good ones. The crowdsourced annotations involve (i)
following the guidelines for \textit{explanation critiquing} to identify applicable critique dimensions of any significant flaws in the explanation, as well as rating the explanation quality; (ii) rating the GPT-4 generated critique.\footnote{See Appendix~\ref{appendix-crowdsource} for more details.} We release expert-revised critiques, crowdsourced explanation scores and flaw dimensions, as well as all critique rating scores as part of \dataset{}.
%\footnote{Worker ids have been anonymized and released.}  
%This also includes annotations on our dev set (DSCB-dev-crowd-anno) which we describe further in Section \ref{sec:experiments-critiquers}.

As part of our evaluation, we also crowdsourced annotations of dev set critiques generated by
our trained critique models (described below in Section~\ref{sec:our-model}).
These annotations are also included in the dataset release, and described
in Section~\ref{sec:experiments-critiquers}.

\eat{
\edit{%To measure the quality of the critique models, 
In addition, we also obtain human annotations on a subset of the \dataset{} dev set (DSCB-dev-crowd-anno, subset) for evaluation purposes. We sample 270 questions across all 10 datasets (50 from each ARC dataset, 20 from each RAINBOW dataset, and 25 each from OBQA and CommonsenseQA).
%MOVED to Appendix
%Since a large fraction of student model explanations receive a ``None'' critique from all 3 critique models (57\%), and this is less interesting to obtain annotations for, we keep only two such instances for each dataset, while randomly sampling instances for the other cases making sure there is at least one instance for every flaw dimension found by the GPT-4 critique model.
Each of the resulting 270 questions with explanations from 4 student models in different styles, along with critiques by 3 critique models for each explanation, were annotated by 3 crowdworkers.\footnote{More sampling and annotation details in Appendix \ref{appendix-details-dataset}.} We use this data to evaluate each critique model on flaw dimensions identified, generated explanation score, and overall critique quality (Sections \ref{subsec:results-gpt4} and \ref{subsec:results-DS-models}).}
}

\begin{table} [t]
\centering
{\small
\begin{tabular}{cl} 
\toprule
\textbf{\makecell[c]{Critique score}} & \textbf{\makecell[l]{Description of critique quality}} \\ \midrule
0 & Bad quality critique \\ \midrule
1 & \makecell[l]{Mostly bad quality critique, but\\makes some useful point(s)} \\ \midrule
2 & \makecell[l]{Mostly good quality, helpful \\
critique, but can be improved} \\ \midrule
3 & \makecell[l]{Very good quality critique} \\
\bottomrule
\end{tabular}
}
\caption{Descriptions for different critique quality scores. Generated critiques are scored on a scale of 0 to 3, where 0 indicates the critique is very bad and 3 means that the critique is of very good quality. To encourage the annotators to make an overall decision about acceptability (2+) of the critique, we avoided including an ``in-between'' category (e.g., 1.5).} \label{tab:critique_score_0-3} 
\vspace{-4mm}
\end{table}

% \subsection{Manual Annotation}

% %Despite careful prompting following the task formulation in Section \ref{sec:explanation_critiquing}, 
% With chosen number of questions * dataset * models * explanation style, obtain critiques generated by GPT-4 as seed data using current prompt  [will oversample questions where models go wrong]

\subsection{Statistics}
%\todo{table with dataset partition, \% total, 0, 1, 2, 3}
%\todo{} \% high quality critiques, \todo{} acceptable, \todo{} flawed, \todo{} very bad
% We introduce \dataset{}, a high-quality data set of around 13,000 entries of such nuanced explanation critiques for around 4,000 questions in the science and commonsense domains. 

% We present a sizeable dataset, \dataset{}, under the \textit{explanation critiquing} 
% task formulation (Section \ref{sec:explanation_critiquing}) as a resource for future efforts in evaluating model-generated reasoning chains. 
%\dataset{} is a diverse dataset of 
The final dataset contains 26,478 entries of nuanced explanation critiques for 4,091 different questions in the science and commonsense domains. %(see Table \ref{tab:dataset_composition} for more details of data composition). 
Table \ref{tab:dataset_composition_simple} summarizes the main statistics of our \dataset{}, with more details in Table~\ref{tab:dataset_composition} in Appendix~\ref{appendix-details-dataset}.
%showing how the partitions cover questions from QA datasets. 
To the best of our knowledge, this is the first dataset of its kind on \textit{explanation critiquing}, covering
nuanced and interpretable (user comprehensible) critiques on different models' explanations and in different explanation styles.
Its roles are to both train automated critique models, and to compare new models' explanation capabilities against.

\section{Our Model: \model{} \label{sec:our-model}}
In addition to probing GPT-4 (not fine-tuned) as a critique model, we also train two smaller models, \modelabb-7B and \modelabb-13B,\footnote{We use \modelabb{} as the short form for \textbf{D}igitial \textbf{S}ocrates.} to explore whether \textit{smaller} language models can be fine-tuned to perform the task of \textit{explanation critiquing} well.

\subsection{Training Data}
\label{subsec:training_data}
We use the train partitions of \dataset{} where we apply two filters: (i) down-sample critiques finding no flaws so they make up at most 50\%; (ii) only keep critiques that are judged as good quality by annotators (with a score of 2 or 3).

We order the training set by degree of annotation, arriving at the following curriculum of instances:
\begin{itemize}[noitemsep, topsep=0pt]
  \item 2032 from DSCB-train-silver (1 epoch)
  \item 1640 from DSCB-train-crowd-anno (1 epoch)
  \item 198 from DSCB-train-expert (2 epochs)
\end{itemize}

\subsection{Training}
We fine-tune two critique models DS-7B and DS-13B starting from Llama2-7B-chat and Llama2-13B-chat \citep{touvron2023llama} respectively. First, we pre-fine-tune on a set of about 50k training questions from ARC and RAINBOW, doing a simple zero-shot question-answering task. Then we further fine-tune on the curriculum of increasing critique quality as indicated above (Section \ref{subsec:training_data}). See Appendix~\ref{appendix-training-hyperparams} for more details.

\section{Experiments}
\label{sec:experiments-critiquers}
We performed experiments and analyses to answer three questions: (1) What types of insights can the {\it explanation critiquing task} reveal about student models? % To do this, we use human-annotated critiques 
(2) How good is GPT-4 at {\it automatic critiquing}? (3) How well can a {\it smaller, trained model} (thus avoiding expensive API calls) perform automated critiquing?

To answer the first question, we analyze the human-annotated training data to
identify what these ``ideal'' critiques can tell us about the student models
(Section~\ref{subsec:result-task-design}).
To answer the second and third questions (i.e., can we generate such critiques automatically?), 
we first ran our three critique models (GPT-4, and the trained
models DS-7B, DS-13B) to generate critiques of explanations in the dev partition of our
dataset (DSCB-dev-crowd-anno). We then obtained human annotations
on a subset of these, namely the three critiques on student model explanations for 270 dev questions
(drawn from all 10 datasets in dev), i.e., 810 annotated critiques (each done by 3 annotators).\footnote{A total of 32 annotators contributed to the whole annotation effort. Further details are in Appendix \ref{appendix-details-dataset}.}
% The annotated dev data is also included in the dataset release.}
Finally we use this data to evaluate each critique model (Sections \ref{subsec:results-gpt4}
and \ref{subsec:results-DS-models}).

\section{Results and Analysis}
\label{results_analysis}
% https://docs.google.com/spreadsheets/d/1J-kiHAmxHZ5fCXXxBz5cIzAQ1I4B4IJnBGRebTTmrzE/edit#gid=0
% For now we focus on with gold
\subsection{What can \textit{explanation critiquing} inform us about student models?}
\label{subsec:result-task-design}
% \textbf{Insights from human annotations}
% Turked 3000, Expert 200, Eval 3000

%The \textbf{explanation critiquing} task
% EXPLANATION SCORE
In this section, we discuss the types of insights we can get from the \textit{explanation critiquing} task, based on human-annotated critique data from the DSCB-train-crowd-anno partition of  \dataset{}. In subsequent sections, we show that this carries over to model-generated critiques as well.

\begin{figure}[t]
\centering
      \includegraphics[width=0.5\columnwidth]{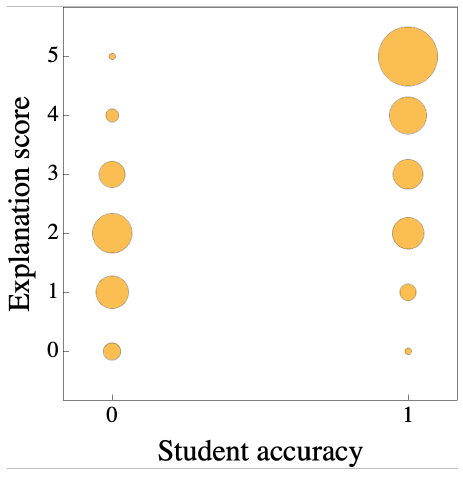}
\caption{In student models, (human-annotated) explanation scores $E_{SC}$ vary greatly within cases where models get the answer right (accuracy = 1) or wrong (accuracy = 0). Even when a model gets the answer correct, its reasoning chain can contain varying degrees of flaws. On the other hand, when a model is incorrect in its answer, it could still make some valid points.}
\label{fig:explanation_score_against_qa_acc_agg}
\vspace{-4mm}
\end{figure}

We analyze the distribution of explanation scores $E_{SC}$ separately for questions where the student model answers correctly vs incorrectly (Figure \ref{fig:explanation_score_against_qa_acc_agg}). Even when a model gets the answer correct, it can give a flawed reasoning chain, ranging from minor flaws ($E_{SC} = 4$) to something completely wrong ($E_{SC} = 0$). Conversely, when a model is incorrect in its answer, it could still make some valid points ($E_{SC} = 2$), thus not all incorrect final answers should be viewed the same as a complete lack of understanding required for the question. Such trends are \textbf{seen across models}, regardless of model size and type (see Figure \ref{fig:explanation_score_against_qa_acc_indiv} in Appendix \ref{appendix-expl-score}).

\begin{figure}[t]
\centering
      \includegraphics[width=\columnwidth]{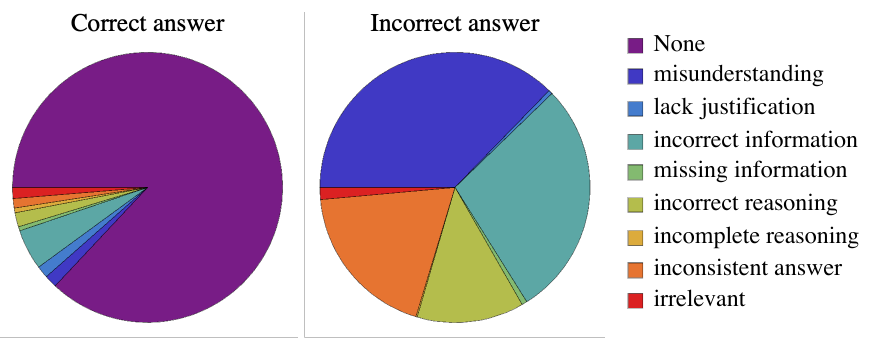}
\caption{The pie charts show distributions of explanation flaws across all student models. Even when models get the answer correct, they may still make errors in their reasoning chain (left). When models answer incorrectly, \textit{explanation critiquing} helps in categorizing and diagnosing errors in the reasoning chain (right).}
\label{fig:flaw_dimensions_correct_incorrect}
\vspace{-4mm}
\end{figure}

% EXPLANATION FLAW DIMENSION
%In addition to explanation score being a useful analysis metric, 
The \textit{explanation critiquing} task also provides the dimension of the most significant flaw, giving much explainability and interpretability. Using GPT-4 generated critiques rated as good quality by crowdworkers (critique score 2 or 3), Figure~\ref{fig:flaw_dimensions_correct_incorrect} summarizes the overall fractions of different significant flaws in models' reasoning chains. When models answer questions correctly, 
%there are many good explanations (no flaws), but 
there can still be cases where the reasoning chain has flaws (13\% across all student models) such as incorrect information. When the model is incorrect, there is a broad spectrum of flaws, such as 
misunderstanding the question or answer options, incorrect information or reasoning, and reasoning inconsistent with the given answer.

\begin{figure}[ht]
\centering
      \includegraphics[width=\columnwidth]{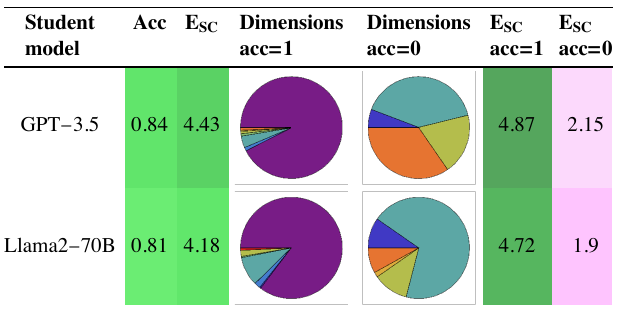}
\caption{GPT-3.5 and Llama-2-70B student models achieve comparable $Acc$ on Science datasets, with the latter having slightly lower $E_{SC}$. They also show different patterns in their explanations flaws, e.g., in the amount of incorrect information vs inconsistent answer.}
\label{fig:flaw_dimensions_compare2}
\vspace{-3mm}
\end{figure}

As a case study, we compare two 
student models, GPT-3.5 and Llama2-70B, on the Science domain
%subset of the data 
(Figure~\ref{fig:flaw_dimensions_compare2}). The models have comparable accuracy, with a slightly lower explanation score $E_{sc}$ for the latter, hinting at shallower understanding. The pie charts showing the flaw dimensions provide an efficient snapshot of the models' weaknesses. For incorrect answers, we see that Llama2-70B has a higher rate of incorrect information, which could potentially be mitigated by information retrieval or otherwise improving factuality. GPT-3.5 instead has a notably large fraction of inconsistent answer, where the answer does not match the reasoning in the explanation. Future work improving GPT-3.5 could potentially experiment with self-reflection, asking if the answer follows from the explanation. We extend this picture across all student models and datasets, discussing further model and domain-specific insights in Appendix \ref{appendix-expl-flaw-dimensions}

%We leave such interesting hypotheses for future work.

Our \textit{explanation critiquing} task formulation delivers even more nuance and interpretability by localizing the flaws. %identified
$f_{loc}$ pinpoints flawed parts of model-generated explanations, like ``Ferromagnetic (C) materials are not typically used for paper clips.''\footnote{Example of incorrect information in Llama2-70B. More examples of identified flaws are in Table~\ref{tab:flaw_examples_gpt4_turker_approved} in Appendix \ref{appendix-expl-flaw-loc-suggestions}.} Beyond looking at the dimensions of explanation flaws to note that incorrect information is frequent in Llama2-70B model's explanations compared to GPT-3.5, localizing the flaws further informs us about topics in which the model has incorrect information. Having a \textbf{list of such flaws} for each model could better \textbf{facilitate finding patterns in models' gaps in knowledge} and reasoning. 

\begin{table*} [t]
\centering 
{\small
\begin{tabularx}{\textwidth}{llX}
\hline
\textbf{\makecell[l]{Student \\model}} & 
\textbf{\makecell[l]{Dataset}} & 
\textbf{\makecell[l]{General feedback examples (acc=0) }} \\
\hline
GPT-4 & Science & 
{\raggedright Remember that the state of matter is indeed a physical property that can distinguish between different types of elements. Metals are generally solid at room temperature, while non-metals can be gases, liquids, or solids.}\\
GPT-3.5 & Science & 
{\raggedright Remember that metals are excellent conductors of electricity due to their free electrons.}\\
Llama2-70B & Science & 
{\raggedright Always remember that in multicellular organisms, the hierarchy of complexity begins with the cell, then progresses to tissues, organs, and finally organ systems.}\\
Llama2-7B & Science & 
{\raggedright When conducting an experiment, it's crucial to control for as many variables as possible to ensure that any observed differences are due to the factor you're testing (in this case, the type of plant food), not other factors.}\\
\hline
GPT-4 & Commonsense & 
{\raggedright Always ensure you understand the question and the answer choices. The question is not asking what Skylar will definitely do next, but what someone might want Skylar to do next.}\\
GPT-3.5 & Commonsense & 
{\raggedright Ensure that your explanation aligns with the answer you have chosen. It's important to be consistent in your reasoning and conclusion.}\\
Llama2-70B & Commonsense & 
{\raggedright Be careful with interpreting the language used in the question and answer choices. Sometimes, phrases may not be literal but idiomatic or figurative.}\\
Llama2-7B & Commonsense & 
{\raggedright Sensitivity to spice typically means a person can't handle much spice, not that they add more spice to their food.}\\
\hline
\end{tabularx}
}
\caption{Examples of general suggestions sampled from human-verified critiques in DSCB-train-crowd-anno. The \textit{explanation critiquing} task includes providing a suggestion that addresses general misconceptions or confusion underlying mistakes that models make under each dimension.}
\label{tab:suggestions_examples_gpt4_turker_approved}
\end{table*}

The \textit{explanation critiquing} task also provides general and specific suggestions to correct the flaw. For instance, providing correct information like ``Ferromagnetic materials, such as iron, nickel, and cobalt, ... in a variety of applications, including holding sheets of paper together.''\footnote{Example of a general suggestion to Llama2-70B's explanation.} The \textbf{general feedback} (Table~\ref{tab:suggestions_examples_gpt4_turker_approved}) could then offer \textbf{directions toward streamlining model improvement} or serve as \textbf{useful retrieval corpus}, while the specific feedback helps to correct reasoning for each instance.

%e.g.,``Ferromagnetic materials, such as iron, nickel, and cobalt, ... in a variety of applications, including holding sheets of paper together.''
%For instance, we observed above that incorrect information is frequent in Llama2-70B model's explanations compared to GPT-3.5. Localizing the flaws further informs us about topics in which the model has incorrect information, e.g., the incorrect "Ferromagnetic (C) materials are not typically used for paper clips" (since paper clips are often made of steel, which is ferromagnetic). Having a \textbf{list of such flaws} for each model could better \textbf{facilitate finding patterns in models' gaps in knowledge} and reasoning. 
%The \textbf{general feedback} (Table~\ref{tab:suggestions_examples_gpt4_turker_approved}) could then provide \textbf{directions towards streamlining model improvement} or \textbf{useful retrieval corpus}, while the specific feedback helps to correct reasoning for each instance.

With  \textit{explanation critiquing}, as informed by the principles of Socratic questioning, the nuanced and interpretable critiques offer various components useful as a systematic, rigorous approach for evaluating model-generated explanations. Such fine-grained information allows for effective uncovering of specific errors associated with models, which can also be used to improve models in the future. 
%(by targeted augmented training, retrieving from previous general feedback, etc). 

\begin{table}[t]
\centering
      \includegraphics[width=\linewidth]{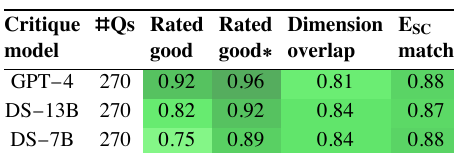}
\caption{Comparison of critique quality from GPT-4, \modelabb-7B, and \modelabb-13B, as rated by crowdworkers. We compare critiques generated by each critique model in terms of the percentage that is rated good (critique score 2 or 3) by workers, how often the flaw dimension identified overlaps with human annotations, and explanation score ($E_{SC}$, cases where model-generated and human-rated explanation score is within 1). The ``Rated good*'' column extrapolates to the actual distribution of the full dataset by accounting for the large fraction (57\%) of examples with None critiques from all 3 models.}
\label{tab:GPT4_versus_DS_turked}
\vspace{-4mm}
\end{table}

%\subsection{To what extent can GPT-4's \textit{explanation critiques} inform us about student models' capabilities?}
\subsection{To what extent can GPT-4 perform the task of \textit{explanation critiquing}?}
\label{subsec:results-gpt4}

While %the different parts of 
\textit{explanation critiquing} allows for systematic, insightful analysis of 
%student 
models, 
%beyond using QA accuracy (see Section \ref{subsec:result-task-design})
obtaining human-corrected or even human-verified explanation critiques can be expensive and time-consuming.
%In this section, 
We analyze if raw critiques generated by GPT-4 can be used to reveal the same trends,\footnote{Compared to obtaining human annotations, API calls are still relatively cheaper in time and monetary cost.} %We perform experiments and analysis on 
using the \dataset{} dev partition (DSCB-dev-crowd-anno).
%which includes questions from two unseen datasets (OBQA and CSQA). 

% annotated sample
%% ROMOVE BELOW: mentioned the table later, moved others to appendix (linked to 6.2)
%Table~\ref{tab:GPT4_versus_DS_turked} summarizes the quality of critiques from different models as rated by crowdworkers. Each critique is rated by 3 workers, and we compute the rounded average of their ratings.
\textbf{Human judgments of critique quality} (Table~\ref{tab:GPT4_versus_DS_turked}) show that 92\% of critiques generated by GPT-4 are rated mostly good or very good quality. This number extrapolates to 96\% when we account for the fact that a large portion of the full dev set consists of explanations that received a ``None'' critique (such instances all rated correct in crowdsourcing) from all 3 critique models. %This number increases to 96\% when we account for the fact that a large portion of the full dev set consists of explanations that received a ``None'' critique from all 3 critique models (GPT-4, DS-13B, and DS-7B), assuming these are all correct (as suggested by the crowdsourcing). We purposefully undersampled these in the data for annotation (7\%) while they are prevalent in full distribution (57\%). 
GPT-4 critiques also \textbf{agree well with crowdsourced \textit{explanation critiquing}}. In 81\% of the cases, the GPT-4 critique is along a dimension that overlaps with annotators' identified flaw dimensions, and 88\% of GPT-4 critiques give an explanation score within 1 of that given by annotators on the 6-point scale (Table~\ref{tab:explanation_score_0-5}), thus providing similar signals in analysis. We also observe very high correlation between annotators and GPT-4 for explanations scores, with Pearson's correlation coefficient $r=0.73$.
%which on the 5-point scale means they are close enough 
%for analysis 
%to provide similar signals in analysis. 
These metrics suggest that analysis done directly using the GPT-4 generated critiques strongly agrees with human-verified data.
%These metrics all suggest that directly using the GPT-4 generated critiques for analysis should have very high correlation to analysis done with human-verified data.% edits and verification.

% We will discuss further examples of insights from GPT-4 generated critiques in the next section, along with critiques generated from our trained Digital Socrates models.

\eat{
% all dev
Using all data points in our dev partition (DSCB-dev-crowd-anno), hen visualizing the student models' flaw dimensions, we also observe that GPT-4 critiques as verified by human annotators reveal comparable trends compared to using raw GPT-4 critiques (see Figure \todo{figure of pie charts comparison for this: gpt-4 on dev rated good by humans vs gpt-4 dev}).

\todo{some examples here??? maybe a big table, 2 whole examples from each model}
\textbf{\textit{explanation critiquing} automatically surfaces model flaws} (e.g. list of incorrect facts)

\textbf{\textit{explanation critiquing} automatically surfaces ways to improve the student models} (e.g. suggestions)
}

\subsection{To what extent can our smaller, trained models (\model{} 7B, 13B) perform the task of \textit{explanation critiquing}?}
\label{subsec:results-DS-models}

%We go one step further to investigate how well our smaller, trained models can perform the task of \textit{explanation critiquing}, moving away from expensive API calls to a proprietary model. As discussed above for GPT-4, we also 
We evaluate the critique quality of \modelabb-7B and \modelabb-13B, both in terms of overall critique quality, as well as flaw dimension overlap and explanation score match compared to human annotations.

% Annotated
Based on \textbf{human judgments of critique quality}  (Table \ref{tab:GPT4_versus_DS_turked}), the vast majority of critiques generated by our \modelabb{} models are rated good or very good, at 82\% for \modelabb-13B and 75\% for \modelabb-7B. These numbers increase to 92\% and 89\% when extrapolating to the full distribution.\footnote{Despite these models being orders of magnitude smaller than GPT-4.} 
%(as discussed above)
%We also observe that 
Our \modelabb{} models' critiques \textbf{highly agree with human \textit{explanation critiquing}}. For both \modelabb-7B and \modelabb-13B, 84\% of the time, the generated critique is along a dimension that overlaps with annotators' identified flaws, comparing favorably to the 81\% dimension overlap for GPT-4 critiques. This suggests our \modelabb{} models are even more closely aligned to human judgments in terms of identifying the flaw dimension in explanations, potentially due to the effect of human-annotated data in their training. The percentage of critiques that give an explanation score within 1 of that given by annotators, is 88\% for \modelabb-7B ($r=0.76$) and  87\% for \modelabb-13B ($r=0.73$), which is comparable to that for GPT-4 (88\%).

\begin{figure}[!ht]
  {\rule{\columnwidth}{0.4pt}}
  \raggedright
  {\small
  \textbf{Question:} When Dennis operates his lawnmower, he notices the engine makes a lot of noise. He also notices that the engine gets very hot. Which best describes the heat and noise generated from the lawnmower? 
(A) a change in phase 
(B) thermal expansion 
(C) an increase in entropy 
(D) mechanical advantage  \\
\textbf{Explanation (GPT-3.5):} 1) The question states that the lawnmower engine makes a lot of noise. 2) The question states that the lawnmower engine gets very hot. 3) Noise and heat are both forms of energy. 4) The noise and heat generated from the lawnmower are a result of the conversion of energy from the fuel to mechanical energy. \\
\textbf{Answer (GPT-3.5):} (D)  [incorrect, should be (C)]

\begin{tabularx}{\columnwidth}{X}
\toprule
\textbf{DS-13B Critique:}\\
\midrule
The explanation states or suggests the following:\newline
* \textbf{Main flaw} (standalone statement): "The noise and heat generated from the lawnmower are a result of the conversion of energy from the fuel to mechanical energy."\newline
* \textbf{Dimension:} incorrect\_information\newline
Consider these points for revising the explanation:\newline
* \textbf{General:} It's important to understand the difference between the different types of energy. Mechanical energy is the energy of motion, while thermal energy is the energy of heat.\newline
* \textbf{Specific:} In the case of the lawnmower, the noise and heat are not a result of the conversion of energy from the fuel to mechanical energy. The noise is a result of the vibration of the engine, while the heat is a result of the friction and combustion of the fuel.\newline
\textbf{Explanation score:} 2\\
\bottomrule
\end{tabularx}
    
    } %small
  \caption{Sample explanation critique from DS-13B.}
  \label{fig:good-critique-ds13}%
  \vspace{-4mm}
\end{figure}%

\section{Further Analysis on \model{}}
% All dev
Having established the validity of the automatic critique models, especially GPT-4 and \modelabb-13B, we use them to analyze all instances in our dev set.
%, rather than just the crowdsourced subset.

\begin{figure}[!ht]
\centering
      \includegraphics[width=\columnwidth]{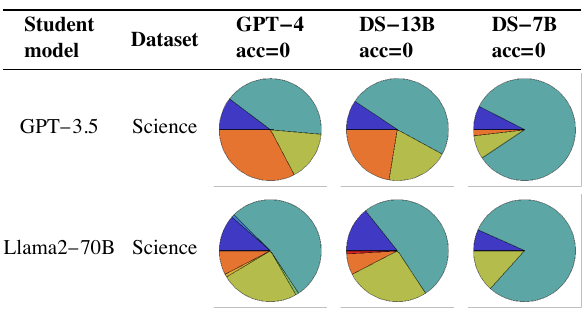}
\caption{Sample comparison of insights on student models based on critiques from GPT-4, \modelabb-7B, and \modelabb-13B using the full dev partition (DSCB-dev-crowd-anno). The distribution of explanation flaws as identified by \modelabb-13B is particularly close to that by GPT-4.}
%when student models got the questions wrong (acc=0).}
\label{fig:GPT4_versus_DS_turked_all_test_small}
\vspace{-4mm}
\end{figure}

\textbf{\modelabb-13B critiques reveal similar trends on student models compared to GPT-4.} 
Figure~\ref{fig:GPT4_versus_DS_turked_all_test_small} presents a representative example comparing the flaw dimension pie charts across critique models. \modelabb-13B and GPT-4 models reveal very similar trends in student models' flaw dimensions, while the \modelabb-7B is slightly less diverse in its critiques.
%and more likely to accept flawed explanations as having no flaws as long as the student correctly answers the question. 
Appendix~\ref{appendix-compare-critics-piecharts} provides the full comparison. % across models and datasets.

%In Figure~\ref{fig:GPT4_versus_DS_turked_all_test_small} we show an example of comparing the flaw dimension pie charts between the critique models. For a full comparison see Figure~\ref{fig:GPT4_versus_DS_turked_all_test_summary} in Appendix~\ref{appendix-compare-critics-piecharts}. We see that the \modelabb-13B and GPT-4 models, in particular, reveal very similar trends in student models' flaw dimensions, while the \modelabb-7B is a little less diverse in its critique, and more likely to accept flawed explanations as having no flaws as long as the student correctly answers the question.

\textbf{Further, \modelabb-13B critiques reveal similar trends on student models compared to human-verified critiques.} Compared to trends identified in Table \ref{tab:flaw_dimensions_correct_incorrect_models_task} using human-verified data, \modelabb-13B identifies similar characterizations of models (Figure~\ref{fig:GPT4_versus_DS_turked_all_test_summary}). For instance, comparing when models answer the questions incorrectly across dataset types, in the case of commonsense reasoning, student models tend to show a misunderstanding of the question or answer choices, whereas incorrect information is more prevalent as a flaw for science datasets. Model-specific insights are also similar. E.g., Llama2-70B's most common flaw when performing scientific reasoning concerns incorrect information. Whereas in GPT-3.5, that flaw is somewhat less frequent but having a reasoning chain inconsistent with the answer is also of concern.
%is of greater concern.

%\todo{1. For the 4 examples on GPT4 vs. DS-13, the point is that they cover the *top common flaw dimensions* -- misunderstanding, None, incorrect\_info, incorrect\_reasoning. And we qualitatively examine that these are all high quality by DS 2. Making use of the generation. For pinpointing flaws, case study on "incorrect\_info" [dump out instances of them from dev -> examples for each model, showing that models have different kinds of incorrect\_info e.g. Llama-7B have more crazy misunderstanding than more subtle than GPT(showing need to go from dimensions to pinpointing); For suggestions, case study on general suggestions -> different models have different things to focus on.}

%\todo{discuss generalization to unseen dataset}

\textbf{\modelabb-13B reveals insightful trends on student models}. 
% explanation score and the distribution of flaw dimensions 
% Quantitatively, in terms of explanation score and the distribution of flaw dimensions identified 
% Table \ref{tab:ds13b_crit_dev_all_datasets_1}
We qualitatively examine flaws localized and suggestions offered by \modelabb-13B. Similar to GPT-4, \modelabb-13B provides high-quality critiques that accurately localize main flaws in models' explanations. Figure~\ref{fig:good-critique-ds13} shows an example of how \modelabb-13B accurately identifies the somewhat subtle error ``The noise and heat generated from the lawnmower are a result of the conversion of energy from the fuel to mechanical energy.'' as incorrect information, providing the accurate feedback of ``... Mechanical energy is the energy of motion, while thermal energy is the energy of heat.''

In Appendix~\ref{appendix-critique-examples}, we cover other top common flaw dimension examples from \modelabb-13B.
%and illustrate that \modelabb-13B is a better performing model compared to the 7B version.
%in Tables~\ref{tab:compare-critiques-gpt4-ds13-p2} (incorrect reasoning) and \ref{tab:compare-critiques-gpt4-ds13-p3} (misunderstanding, None). Table \ref{tab:compare-critiques-ds7-ds13} illustrate how  \modelabb-13B still provides correct and accurate critiques when the smaller DS-7B model is producing invalid critiques.
%In Appendix~\ref{appendix-critique-examples}, we show where the latter table is showing examples of failure cases in which the smaller DS-7B model is producing invalid critiques.
We further demonstrate, in Appendix \ref{appendix-13b-dataset-model},  how comprehensive results for applying DS-13B on all 10 datasets of the \dataset{} dev set, across all 4 student models, reveal a rich diversity of behavior.

\section{Conclusion}

Despite advances in LLMs, the nature and quality of their explanations are still poorly understood.
To address this, we introduced the task of \textit{explanation critiquing} which provides quantitative and qualitative insights into LLM performance, beyond answer accuracy. Our analysis highlights the value of closely examining model-generated explanations %as a window into their reasoning and 
for a deeper understanding of models' capabilities.
We create \dataset{}, a dataset of 26,478 nuanced and interpretable critiques under the \textit{explanation critiquing} task formulation, along with
%partitions containing 
crowdsourced and expert annotations.
%We create \dataset{}, a dataset of student model explanations with critiques under the \textit{explanation critiquing} task formulation. \dataset{} contains 26,478 entries of nuanced and interpretable critiques, along with partitions containing crowdsourced and expert annotations.
Our Digital Socrates model (DS-13B), is a capable open critique model comparable in performance to GPT-4 and closely aligned with human annotators in the \textit{explanation critiquing} task. We also showed how Digital Socrates can, for the first time, provide rich analyses and insights across a range of student models and datasets, without relying on expensive API calls or human annotation.
This work fills an important gap in
evaluation tools for understanding and improving the explanation behavior
of models.

\section*{Acknowledgements}
We thank the anonymous reviewers, as well
as members of the Aristo team at AI2 for their insightful discussions and valuable feedback on an earlier draft.

%\clearpage
%\vspace{10mm}
%\newpage
\section*{Limitations}
%\todo{}

We have demonstrated the effectiveness of \model{} to provide evaluation and insights into a model's capabilities going beyond model accuracy. While exciting, there are limitations in the current work and opportunities for the future.

%The \model{} models were developed and tested based on a varied set of explanation styles, student models, and datasets. This diversity of training should help \model{} generalize to different explanation styles, models or datasets. 
%When applying to diverse downstream use cases, we leave it to future work to further investigate and validate the extent of this generalizability, especially when applied to different domains or task formats. 

Our approach is designed for examining model-generated reasoning chains involving natural language explanations or reasoning steps, focusing on focus on linguistic reasoning (in broad domains of commonsense and science). Building on a better understanding of such reasoning chains, we hope our work can inspire future efforts towards extending the approach in various directions, for instance, for application to longer text e.g., model-generated summaries or stories, as well as symbol manipulation reasoning such as arithmetic, and formal logic. We encourage other researchers to build upon our work in other domains, and even evaluate our existing Digital Socrates models on other reasoning types since the prompt to the models and inherent Socratic questioning principles are domain agnostic. It is thus possible the DS models would apply to such reasoning styles also, and would be an interesting avenue for future work.
%The current datasets focus on linguistic reasoning (in broad domains of commonsense and science), but they do not explicitly cover some other types of reasoning, in particular symbol manipulation reasoning such as arithmetic, and formal logic.
%with real-world implications.

Another interesting direction for future work is investigating how well different student models can respond to the feedback, and to what extent feedback from $s_{gen}$ can generalize to help student models with related questions.

The current \model{} models have to rely on internal knowledge, so like any LLM can be subject to hallucination of factual knowledge. We encourage the use of \model{} models for research purposes only and for researchers to build on them responsibly. 

%Despite the use of parameter-efficient approaches for fine-tuning, the current approach presented still requires the use of GPU resources for model training. 

\section*{Ethics Statement}
%\todo{}
All annotators that participated in the data collection process have been anonymized.  The only personal information we collect is the worker IDs from Amazon
Mechanical Turk, which we will not release. No personally identifiable information is contained in our dataset or otherwise released. We took great care to pay fair
wages, and were responsive to feedback and questions throughout the data collection process. This study involves the use of large-scale language models.
Like any other large-scale
language model, despite the best intentions, there is a risk of our models producing biased or offensive statements as part of the critique text. We release our models for research purposes only.

\bibliography{anthology,custom}

\appendix
\onecolumn

\section{FAQs}
 \begin{itemize}
     \item [\textbf{Q:}] \textbf{What is the role of \dataset{} in future work?}
     \item [] There are three key roles:
     \begin{enumerate}[label=(\alph*)]
    \item Comparative analysis of different models' explanation capabilities: Given a user has a new model to test, they would run our dataset's test questions on their model, then use one of our Digital Socrates to critique the model's answers. It is important to run the same test questions in such tests so that the results can be compared with those generated by other models, if some kind of comparative analysis is needed (e.g., is $GPT_n$ generating better explanations than $GPT_{n-1}$?)
    \item Creation of new Digital Socrates: As new models become available (e.g., new small-but-powerful models on the horizon), the dataset can be used to create new (and hopefully even stronger) Digital Socrates critiques models. The explanations in the dataset were generated by different models including strong proprietary models and open models of different strengths, and thus cover a broad variety of styles/techniques. By training Digital Socrates using this data, our model is designed to be able to robustly evaluate multiple styles of explanation. 
    \item General understanding of explanations: This is the first dataset containing detailed critiques of explanations, and as such could be useful to future researchers wanting to develop better explanation generation methods.
    \end{enumerate}
    
    \item [\textbf{Q:}] \textbf{Is the dataset creation replicable in the future?}
    \item [] Yes! It's important to note that the dataset contains human-vetted GPT-4 critiques (of explanations from 4 different student models), not simply GPT-4 output. Thus the dataset is not a naive reflection of GPT-4, but of human judgments about what critiques are good and which are not - this is critically important. It would be easy to repeat this process using other models, along with vetting to expand the dataset (GPT-4 gave us the highest "hit rate" of good examples). So the process is easily repeatable, just like other AI datasets.
    
    \item [\textbf{Q:}] \textbf{Why are the critiques interpretable?}
    \item [] The intended contrast here is between prior methods offering just a single score ("uninterpretable", in that the reason for the score is opaque), vs. the Digital Socrates, where the critiques tell you the category of the main error (along 8 dimensions guided by Socratic questioning principles), as well as actionable hints to improve the explanation, i.e., user comprehensible ("interpetable") reasons why the explanation is good or bad.
    
    \item [\textbf{Q:}] \textbf{Why is the approach rigorous?}
    \item [] Digital Socrates critiques methodically step through several aspects of a critique (main flaw, dimension, correction, etc.) rather than just giving vague feedback like "The explanation is not detailed enough". Section \ref{sec:explanation_critiquing} introduces the approach and details why it is systematic and methodical.
    
    \item [\textbf{Q:}] \textbf{For rating critique quality, the gap between scores 1 ("mostly bad") and 2 ("mostly good") is rather large?}
    \item [] Yes, it's a balancing act between being too coarse (0/1) and too fine (too many categories for a user to choose from). We could have introduced an "in-between" category (~1.5), but - analogous to reviewing papers - we wanted to encourage the annotators to make an overall decision about acceptability (2+) of the output, rather than just say a less helpful, undirected "in-between" assessment. 
    
    \item [\textbf{Q:}] \textbf{How do you ensure the quality of crowd annotation?}
    \item [] To ensure the quality of our crowdworker annotation, here are some measures we took:
    \begin{enumerate}[label=(\alph*)]
    \item In the design of the task, we ask workers to follow the guidelines for explanation critiquing to identify significant flaws and rate the explanation quality first. This stage is designed to familiarize the workers with the task by trying it out themselves and to let them form their own judgment regarding the quality of the explanation. After which, we then ask them to rate the GPT-4 generated critique. This ensures that their judgment is not just swayed by the generated critique.
    \item In the “rules” section of our instructions to crowdworkers, we noted: \textit{``Finally, note that if you work on this HIT, please answer with care: Some HITs will be checked by hand, and work may be rejected if there are too many errors (e.g., rating a poor critique with a high critique score of 3).''}
    \item During data collection, we first launched a set of 100 questions (along with explanations and critiques) as a pilot trial and sampled the annotations to check that our procedure yielded reasonable responses before launching the full set.
    \item Percentages of critiques rated as good or perfect by crowdworkers (92\%) are comparable to those in expert annotations (86\%).
    \end{enumerate}
    
%    \item [\textbf{Q:}] \textbf{Inter-annotator scores? Critique dimensions can be subjective?}
     \item [\textbf{Q:}] \textbf{How subjective are the critique dimensions? How much do annotators agree?}
    \item [] Indeed, there could a degree of subjectivity in identifying what is the most significant flaw dimension. From our annotation data, quantitatively, we observe that in 91.9\% of the cases, there is at least some overlap between the flaw dimensions identified by the annotators (i.e. at least 2 annotators agree on a dimension). In terms of giving an explanation score, in 89\% of the instances, all annotator's explanation score is within 1 of the mean explanation score for that instance.

\end{itemize}

\section{Details on Crowdsourcing for Estimating Quality of Explanation Critiques}
\label{appendix-crowdsource}

The crowdsourcing instructions included a detailed description of an ideal critique, along with four detailed examples for the different critique quality scores. The workers were asked to first provide their own set of explanation flaw dimensions as well as an explanation score (0-5) before being presented with the model-generated critique to give a quality score (0-3).

\begin{figure}[ht]
\centering
      \includegraphics[width=\columnwidth]{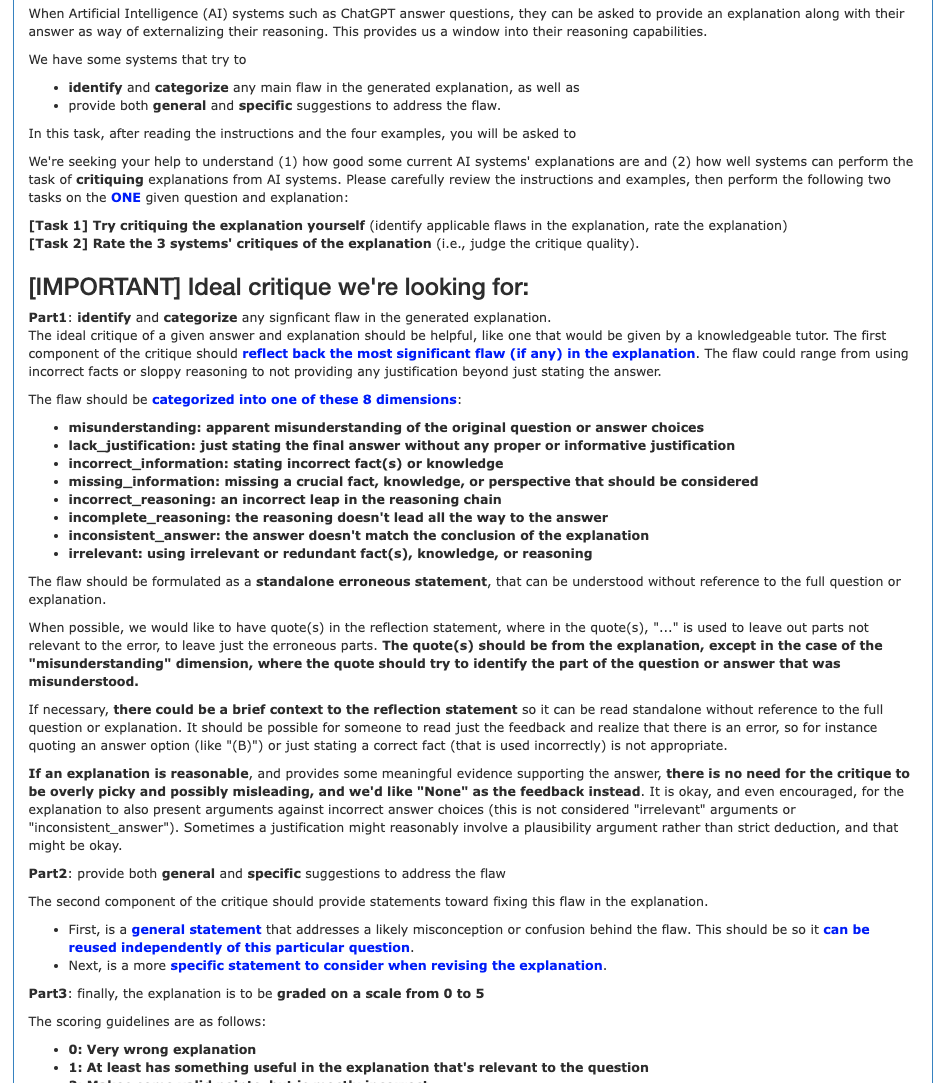}
\caption{Part of the instructions to Amazon Mechanical Turk workers.}
\label{fig:mturk_instructions}
%\vspace{-4mm}
\end{figure}

Figure~\ref{fig:mturk_instructions} shows a screenshot of the initial part of the instructions to the workers.

\noindent \textbf{Turk workers and pay rate:} \\ 
Our participants were recruited on the Amazon Mechanical Turk platform. The workers met minimum
qualification in AMT: 95\% approval rate. They were from US locations and rated at Amazon’s Masters
Level. The workers were paid at a rate
of $\approx$\$15/hr.

\section{Details of Explanation Critiquing task}
\label{appendix-details-task}
\subsection{Details on flaw localization, $f_{loc}$}
\label{appendix-details-task-flaw-localize}
In identifying the most significant flaw, when possible, we encourage quote(s) in this statement to better localize the flaw. In the quote(s), ``...'' is used to leave out parts not relevant to the flaw, and to keep just the erroneous parts. The quote(s) should be from the explanation, except in the case of pointing out a misunderstanding of the question and/or answer options, where the quote should try to identify the part of the question or answer that was misunderstood.

Following the spirit of the types of Socratic questions \citep{Paul2006-PAUTTG-6}, we focus on more significant flaws and ignore more trivial criticisms e.g., upper-case/lower-case issues, small typos, and spacing inconsistencies.

\section{Details of \dataset{} composition}
\label{appendix-details-dataset}

Table~\ref{tab:dataset_composition} shows details of the composition of each partition of \dataset{}. (Table \ref{tab:dataset_composition_simple} summarizes the main statistics of our \dataset{}.)

For science questions, the ARC dataset has both a Challenge Set and an Easy Set, where the Challenge Set contains only questions that are regarded as more difficult (answered incorrectly by retrieval-based and word co-occurence algorithms). 

The \textbf{DSCB-train-crowd-anno} partition was purposefully skewed towards challenging questions, by sampling such that the Llama-7B-chat student model had low 30\% accuracy (vs the natural 60\% accuracy). We favor sampling from Rainbow-Dev because the dev and test partitions of datasets like CosmosQA and Social IQa are designed to be of higher quality. CosmosQA sampled instances from workers who excelled in the generation task for their dev and test, while Social IQa's dev and test were validated by workers a second time. In this partition, each critique is rated by 1 worker. 

In \textbf{DSCB-dev-crowd-anno}, a large fraction (57\%) of student model explanations receive a ``None'' critique from all 3 critique models, and this is less interesting to obtain annotations for. Therefore, in the subset to annotate (DSCB-dev-crowd-anno, subset), we keep only two such instances for each dataset, while randomly sampling instances for the other cases making sure there is at least one instance for every flaw dimension found by the GPT-4 critique model. In this partition, we obtain annotations from 3 workers for each critique and compute the rounded average of their ratings for our analysis.

In total, 26 different crowdworkers contributed to the annotations in DSCB-train-crowd-anno and 22 to annotations in DSCB-dev-crowd-anno. The specific anonymized worker ids have also been included in the released data, giving more details on how the annotations were distributed across workers.

\begin{table*} [t]
\centering 
{\small
\begin{tabular}{cllll} 
\hline
\textbf{\makecell[c]{\dataset{}\\partition}} & 
\textbf{\makecell[c]{Questions from \\QA tasks}} 
& \textbf{\makecell[c]{Explanations\\from student models}} &
\textbf{\makecell[c]{Critique \\instances}} & 
\textbf{\makecell[c]{Human \\annotations}}\\ 
\hline
 \makecell[l]{DSCB-train-silver} & \makecell[l]{3240 questions: \\810 ARC-Challenge-train \\+ 810 ARC-Easy-Train \\+ 270*6 Rainbow-Train} &  \makecell[l]{Each question with \\ONE of \\4 student models \\ * 3 explanation styles} & \makecell[l]{3240 \\Q+A+E+Crit\\from GPT-4} & \makecell[l]{NA} \\ 
\hline 
 \makecell[l]{DSCB-train-crowd-anno} & \makecell[l]{270 questions: \\100 ARC-Challenge-Train \\+ 50 ARC-Easy-Train \\+ 20*6 Rainbow-Dev} &  \makecell[l]{Each question with \\4 student models \\ * 3 explanation styles} & \makecell[l]{3240 \\Q+A+E+Crit\\from GPT-4} & \makecell[l]{crowdsourced\\(all)} \\ 
\hline
\makecell[l]{DSCB-train-expert} & \makecell[l]{31 questions: \\18 ARC-Challenge-Dev \\
+ 13 Rainbow-Dev} &  \makecell[l]{Sampling of \\
4 student models\\
and 3 explanation styles} & \makecell[l]{198\\manually edited \\based on\\GPT-4 output} & \makecell[l]{expert} \\ 
\hline
 \makecell[l]{DSCB-dev-crowd-anno} & \makecell[l]{550 questions: \\100 ARC-Challenge-Test \\+ 50 ARC-Easy-Test \\+ 50*6 Rainbow-Dev \\(no overlap with earlier usage) \\+ 50 OBQA-Test \\+ 50 CSQA-Dev} &  \makecell[l]{Each question with \\4 student models \\ * 3 explanation styles} & \makecell[l]{6600\\
 Q+A+E+Crit\\
 from critiquers:\\
 GPT-4\\DS-7B\\DS-13B
 } & \makecell[l]{crowdsourced\\
 (subset)
} \\
\hline
\end{tabular}
}
\caption{Summary of data composition in \dataset{}. \label{tab:dataset_composition} }
%\vspace{-3mm}
\end{table*}

\section{Explanation Critiquing as a way to go beyond accuracy}
\label{appendix-expl-critiquing-beyond-acc}

\subsection{Examining quality of reasoning chain beyond accuracy}

\begin{figure}[h]
\centering
      \includegraphics[width=\columnwidth]{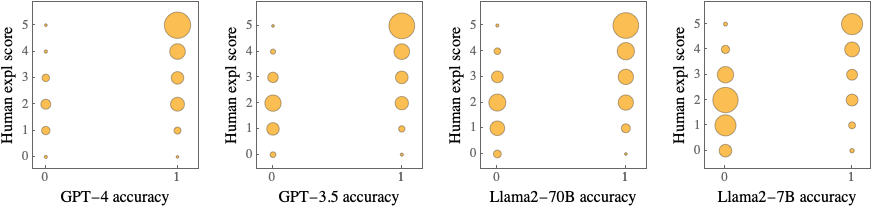}
\caption{Across four student models, human-annotated explanation scores vary greatly within cases where models get the answer right (accuracy = 1) or wrong (accuracy = 0). }
\label{fig:explanation_score_against_qa_acc_indiv}
%\vspace{-4mm}
\end{figure}

In Figure~\ref{fig:explanation_score_against_qa_acc_indiv} we show how human annotated explanation scores vary for correct and incorrect answers across the four student models.
\label{appendix-expl-score}

\subsection{Using explanation flaw dimensions to offer interpretability to explanation scores}
\label{appendix-expl-flaw-dimensions}
% MOVED FROM MAIN TEXT, incorporated below
%We extend this picture across all student models and datasets in Table~\ref{tab:flaw_dimensions_correct_incorrect_models_task} in Appendix \ref{appendix-expl-flaw-dimensions}. We see that even when explanation scores are similar and generally correlate with accuracy, the fine-grained dimensions indicate differences in models’ capabilities and weaknesses. We discuss further examples of model-specific insights in Appendix \ref{appendix-expl-flaw-dimensions}.

Table~\ref{tab:flaw_dimensions_correct_incorrect_models_task}
shows the explanation flaw dimensions across all student models and tasks. We see that even when explanation scores are similar and generally correlate with accuracy, the fine-grained dimensions indicate differences in models’ capabilities and weaknesses. 

In Section~\ref{subsec:result-task-design} we already looked at the case study comparing GPT-3.5 and Llama2-70B student models on the Science datasets. We see from Table~\ref{tab:flaw_dimensions_correct_incorrect_models_task} that the smaller Llama2-7B model shares many of the same patterns as its bigger counterpart but with a larger fraction of ``misunderstanding'' and ``inconsistent answer''. 

Another observation when comparing models' errors in science versus commonsense reasoning is that the flaws in their reasoning chains tend to be due to ``incorrect information'' for science reasoning compared to ``misunderstanding'' of the question or answer options when doing commonsense reasoning. 

Such insights allow us a deeper understanding of models’ capabilities and can be useful for further research to improve model reasoning capabilities in a more targeted manner.

\begin{table}[h]
\centering
      \includegraphics[width=0.8\columnwidth]{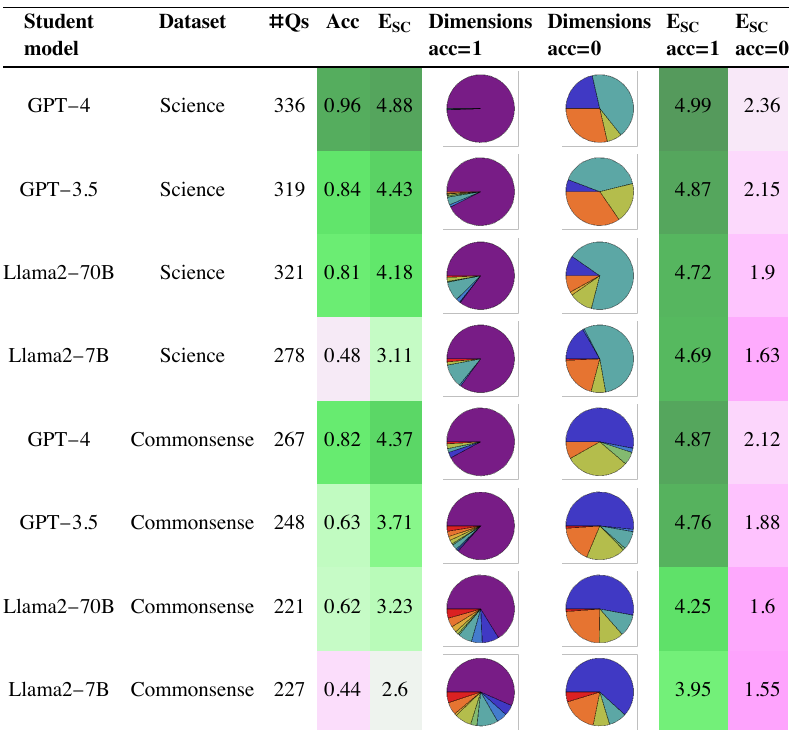}
\caption{Analyzing critiques rated as good by crowdworkers shows that the type of errors in the reasoning chain varies across models and also depends on the type of task dataset. Refer to the legend in Figure \ref{fig:flaw_dimensions_correct_incorrect} for dimensions of flaws. Explanation scores as a summary metric do not capture such nuances and characteristics across models and datasets.}
\label{tab:flaw_dimensions_correct_incorrect_models_task}
%\vspace{-4mm}
\end{table}

\subsection{Allowing for fine-grained analysis through error localization and suggestions}
\label{appendix-expl-flaw-loc-suggestions}
Table~\ref{tab:flaw_examples_gpt4_turker_approved} illustrates more examples of localized flaws across student models and domains. More examples of general suggestions are in Table~\ref{tab:suggestions_examples_gpt4_turker_approved}. Such fine-grained information effectively uncovers specific errors associated with models and offers insights for future work on model improvement.

\subsection{Digital Socrates DS-13B analysis on individual datasets}
\label{appendix-13b-dataset-model}

In Tables~\ref{tab:ds13b_crit_dev_all_datasets_1} and \ref{tab:ds13b_crit_dev_all_datasets_2} we provide the accuracy, explanation scores and flaw dimension pie charts produced by the DS-13B critique models for each of the 10 datasets in the \dataset{} dev set. 

Analyzing these results across all 4 student models, reveals a rich diversity of behavior. E.g., on Physical IQa the models all tend to have more incorrect information than in other RAINBOW datasets, while CosmosQA, especially for GPT-3.5, has a large fraction of incorrect reasoning. Analyzing the individual instances for each dominant explanation flaw will provide more insights into model weaknesses. 

\subsection{Comparing explanation flaws as judged by different critique models}
\label{appendix-compare-critics-piecharts}

\begin{figure*}[!ht]
\centering
      \includegraphics[width=\linewidth]{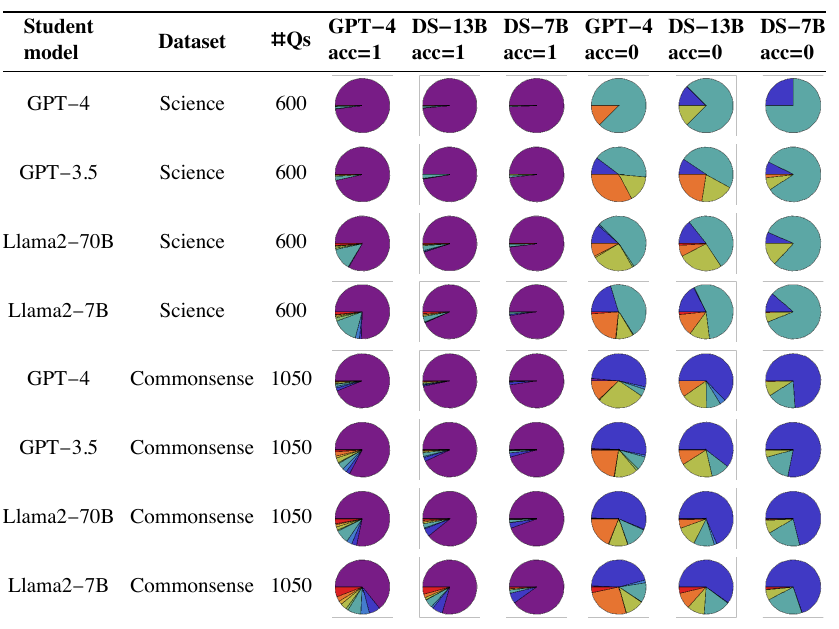}
\caption{Comparison of insights on student models based on critiques from GPT-4, \modelabb-7B, and \modelabb-13B. On all data points in our dev partition (DSCB-dev-crowd-anno), the distribution of explanation flaws as identified by \modelabb-13B is particularly close to that by GPT-4, especially in cases when student models got the questions wrong (acc=0).}
\label{fig:GPT4_versus_DS_turked_all_test_summary}
%\vspace{-4mm}
\end{figure*}

In Figure~\ref{fig:GPT4_versus_DS_turked_all_test_summary}, we show the full comparison of explanation flaw pie charts based on critiques from GPT-4, \modelabb-7B, and \modelabb-13B. We observe that the \modelabb-13B and GPT-4 models, in particular, reveal very similar trends in student models' flaw dimensions, while the \modelabb-7B is a little less diverse in its critique dimensions, and more likely to accept flawed explanations as having no flaws as long as the student correctly answers the question.

% \begin{figure}[h]
% \centering
%       \includegraphics[width=\columnwidth]{images/pie_chart_table_gpt4_filtered_train.pdf}
% \caption{GPT-4 critiques show that the type of errors in the reasoning chain varies across models and also depends on the type of task dataset. Refer to the legend in Figure \ref{fig:flaw_dimensions_correct_incorrect} for dimensions of flaws. Similar trends are observed as when analyzing the subset of critiques rated as good by workers. \todo{make this one a GPT unfiltered version, show that it is similar to filtered}}
% \label{tab:flaw_dimensions_correct_incorrect_models_task_gpt4_unfiltered}
% %\vspace{-4mm}
% \end{figure}

\begin{table*} [t]
\centering 
{\small
\begin{tabularx}{\textwidth}{llX}
\hline
\textbf{\makecell[l]{Student \\model}} & 
\textbf{\makecell[l]{Dataset}} & 
\textbf{\makecell[l]{Dimension (acc=0) with frequency and example}} \\
\hline
GPT-4 & Science & 
{\raggedright {\textbf{incorrect information} (0.88): "The Sun, which is a sulfur burning heavenly body."}}\\
 &  & 
{\raggedright {\textbf{inconsistent answer} (0.12): "Therefore, the most likely outcome is a decrease in the population of blue flowers."}}\\
GPT-3.5 & Science & 
{\raggedright {\textbf{incorrect information} (0.41): "Option (D) is incorrect because it incorrectly states that birds must eat food to maintain a constant temperature, when in fact they can regulate their body temperature internally."}}\\
 &  & 
{\raggedright {\textbf{inconsistent answer} (0.33): "Given answer: (C)" while the explanation supports answer (D).}}\\
 &  & 
{\raggedright {\textbf{incorrect reasoning} (0.16): "Longer spines caused a genetic change in the fish."}}\\
 &  & 
{\raggedright {\textbf{misunderstanding} (0.1): "Thermometers indicate levels of mercury in the blood."}}\\
Llama2-70B & Science & 
{\raggedright {\textbf{incorrect information} (0.53): "Ferromagnetic (C) materials are not typically used for paper clips."}}\\
 &  & 
{\raggedright {\textbf{incorrect reasoning} (0.25): "A decrease in the length of the ramp would result in a shorter distance over which the force must be applied, thereby requiring less force."}}\\
 &  & 
{\raggedright {\textbf{misunderstanding} (0.11): "Mutualism is a relationship in which both organisms benefit."}}\\
 &  & 
{\raggedright {\textbf{inconsistent answer} (0.08): "Given answer: (B)" while the explanation correctly describes clear and sunny weather, it incorrectly chooses option B as the correct answer.}}\\
 &  & 
{\raggedright {\textbf{incomplete reasoning} (0.01): }}\\
Llama2-7B & Science & 
{\raggedright {\textbf{incorrect information} (0.46): "Longer spines caused a genetic change in the fish."}}\\
 &  & 
{\raggedright {\textbf{inconsistent answer} (0.22): "If the load is reduced by half, the force required to move the cart remains the same, so the cart will move faster (B)."}}\\
 &  & 
{\raggedright {\textbf{misunderstanding} (0.2): "Water changing from liquid to solid is described by option (C) snow melting into a puddle."}}\\
 &  & 
{\raggedright {\textbf{incorrect reasoning} (0.1): "Many of these bacteria became extinct due to changes in the Earth's environment that led to an increase in oxygen levels."}}\\
 &  & 
{\raggedright {\textbf{irrelevant} (0.01): "A mammal is a warm-blooded animal that has hair, produces milk for its young, and has a specific type of nervous system."}}\\
\hline
GPT-4 & Commonsense & 
{\raggedright {\textbf{misunderstanding} (0.52): "Without additional information, we cannot determine the speaker's preference."}}\\
 &  & 
{\raggedright {\textbf{incorrect reasoning} (0.28): "The location where he preferred to play and where he sounded quieter must be the same."}}\\
 &  & 
{\raggedright {\textbf{inconsistent answer} (0.12): "Given answer: (C)" while the explanation suggests that Bailey would feel cautious of things around her (B).}}\\
 &  & 
{\raggedright {\textbf{incorrect information} (0.04): "Smoke can be seen in the air due to the light scattering off particles, but it does not color the air itself."}}\\
 &  & 
{\raggedright {\textbf{lack justification} (0.01): "However, the context does not provide specific information about Jan's preferences or intentions, so any of the options could be possible depending on Jan's personality and mood."}}\\
GPT-3.5 & Commonsense & 
{\raggedright {\textbf{misunderstanding} (0.53): "The logical continuation would be that a final spot free rinse (A) is taken from under the hood."}}\\
 &  & 
{\raggedright {\textbf{inconsistent answer} (0.22): "Therefore, 'mortify' (B) is the best answer choice that aligns with the context of the sentence."}}\\
 &  & 
{\raggedright {\textbf{incorrect reasoning} (0.14): "Jordan would feel 'over' the relationship, indicating that they are ready to move on."}}\\
 &  & 
{\raggedright {\textbf{incorrect information} (0.08): "The cooking method that is more suitable for poor weather conditions is smoking, as it can be done indoors."}}\\
 &  & 
{\raggedright {\textbf{missing information} (0.01): "The context does not mention any reason for Natasha to go back inside the villa in the evening."}}\\
Llama2-70B & Commonsense & 
{\raggedright {\textbf{misunderstanding} (0.56): "Quinn needs to work before he can afford nice meals."}}\\
 &  & 
{\raggedright {\textbf{inconsistent answer} (0.19): "Building a special room with no windows and no tech type of lights can create a sleep-conducive environment."}}\\
 &  & 
{\raggedright {\textbf{incorrect information} (0.13): "A stainless steel bucket (A) cannot be used to milk a cow into because it is not designed for that purpose."}}\\
 &  & 
{\raggedright {\textbf{incorrect reasoning} (0.11): "The writer asked if she should bring a swimsuit because it's good to wear for playing Scattergories Junior."}}\\
 &  & 
{\raggedright {\textbf{lack justification} (0.01): "The correct answer will describe the steps to open a bolt using a tool."}}\\
Llama2-7B & Commonsense & 
{\raggedright {\textbf{misunderstanding} (0.45): "The sentence implies that Katrina had very short hair."}}\\
 &  & 
{\raggedright {\textbf{inconsistent answer} (0.26): "The context mentions that he was speaking at a mile a minute, which could cause him to not make sense."}}\\
 &  & 
{\raggedright {\textbf{incorrect information} (0.12): "A stainless steel bucket can be used to insulate against boiling water."}}\\
 &  & 
{\raggedright {\textbf{incorrect reasoning} (0.11): "This suggests that Natalie did not have a significant other at the time of the dance. Maria, on the other hand, did have a significant other, as she was able to leave the dance to get food."}}\\
 &  & 
{\raggedright {\textbf{irrelevant} (0.03): "To keep a plant healthy, it is important to provide it with adequate water, nutrients, and light."}}\\

\hline
\end{tabularx}
}
\caption{Examples of localized flaws sampled from human-verified critiques in DSCB-train-crowd-anno. The \textit{explanation critiquing} task pinpoints the specific mistakes that models make under each dimension.}
\label{tab:flaw_examples_gpt4_turker_approved}
\end{table*}

\section{More critique examples}
\label{appendix-critique-examples}
More examples from \modelabb-13B are in Figures~\ref{fig:compare-critiques-gpt4-ds13-p2} (incorrect reasoning, incorrect information) and \ref{fig:compare-critiques-gpt4-ds13-p3} (misunderstanding, None), covering other common flaw dimensions. These further examples of explanation critiques from GPT-4 vs DS-13B illustrate that, like GPT-4, DS-13B also provides high quality explanation critiques across different dimensions. Table \ref{tab:compare-critiques-ds7-ds13} illustrates how \modelabb-13B still provides correct and accurate critiques when the smaller DS-7B model is producing invalid critiques, suggesting that the former is better performing.

%Tables~\ref{tab:compare-critiques-gpt4-ds13-p2} and \ref{tab:compare-critiques-gpt4-ds13-p3}  shows more examples of explanation critiques from GPT-4 vs DS-13B. Table~\ref{tab:compare-critiques-ds7-ds13} shows examples of failure cases in which DS-7B produces low quality critiques compared to DS-13B.

\begin{figure*}[!htbp]
  {\rule{\textwidth}{0.4pt}}
  \raggedright
  {\small
  
  \textbf{Question:} A student is pushing a 20-kilogram box up a ramp. Which change will require the student to use less force to push the box? 
(A) increasing the mass of the box 
(B) decreasing the length of the ramp 
(C) decreasing the steepness of the ramp 
(D) increasing the friction on the surface of the box \\
\textbf{Explanation (Llama2-70B):} 1) The force required to push a box up a ramp is determined by the weight of the box and the angle of the ramp.\linebreak
2) A decrease in the length of the ramp would result in a shorter distance over which the force must be applied, thereby requiring less force.\\
\textbf{Answer (Llama2-70B):} (B)  [incorrect, should be (C)]

    \begin{tabularx}{\linewidth}{XX}
    \\
    \toprule
    Critique GPT-4 & Critique DS-13B\\
    \midrule
The explanation states or suggests the following:\newline
 * Main flaw (standalone statement): "A decrease in the length of the ramp would result in a shorter distance over which the force must be applied, thereby requiring less force."\newline
 * Dimension: incorrect\_reasoning\newline

Consider these points for revising the explanation:\newline
 * General: The force required to move an object up a ramp is not directly related to the length of the ramp, but rather to the steepness of the ramp and the weight of the object. Shortening the ramp does not decrease the force needed, it just decreases the distance over which the force is applied.\newline
 * Specific: In this case, the correct answer is (C) decreasing the steepness of the ramp. This would decrease the vertical component of the force required to push the box up the ramp, thereby requiring less force overall.\newline

Explanation score: 2
&
The explanation states or suggests the following:\newline
* Main flaw (standalone statement): "A decrease in the length of the ramp would result in a shorter distance over which the force must be applied, thereby requiring less force."\newline
* Dimension: incorrect\_reasoning\newline

Consider these points for revising the explanation:\newline
* General: Remember that the force required to push an object up a ramp is determined by the weight of the object and the angle of the ramp, not the length of the ramp.\newline
* Specific: In this case, the force required to push the box up the ramp is determined by the weight of the box and the angle of the ramp, not the length of the ramp. Therefore, a decrease in the length of the ramp would not necessarily require less force.\newline

Explanation score: 2\\
    \bottomrule
    \\
    \end{tabularx}%

  \textbf{Question:} When Dennis operates his lawnmower, he notices the engine makes a lot of noise. He also notices that the engine gets very hot. Which best describes the heat and noise generated from the lawnmower? 
(A) a change in phase 
(B) thermal expansion 
(C) an increase in entropy 
(D) mechanical advantage  \\
\textbf{Explanation (GPT-3.5):} 1) The question states that the lawnmower engine makes a lot of noise.\newline
2) The question states that the lawnmower engine gets very hot.\newline
3) Noise and heat are both forms of energy.\newline
4) The noise and heat generated from the lawnmower are a result of the conversion of energy from the fuel to mechanical energy. \\
\textbf{Answer (GPT-3.5):} (D)  [incorrect]

    \begin{tabularx}{\linewidth}{XX}
    \\
    \toprule
    Critique GPT-4 & Critique DS-13B\\
    \midrule
The explanation states or suggests the following:\newline
 * Main flaw (standalone statement): "The noise and heat generated from the lawnmower are a result of the conversion of energy from the fuel to mechanical energy."\newline
 * Dimension: incorrect\_information\newline

Consider these points for revising the explanation:\newline
 * General: It's important to understand that while noise and heat are indeed forms of energy, they are often considered waste or unwanted byproducts in many mechanical systems, not the desired output. This is because they represent energy that is not being efficiently converted into the intended form (like mechanical work), and instead is being dissipated in less useful ways.\newline
 * Specific: In the context of this question, the noise and heat generated by the lawnmower engine are not the result of the conversion of energy to mechanical advantage, but rather represent an increase in entropy, which is a measure of disorder or randomness in a system. \newline

Explanation score: 2
&
The explanation states or suggests the following:\newline
* Main flaw (standalone statement): "The noise and heat generated from the lawnmower are a result of the conversion of energy from the fuel to mechanical energy."\newline
* Dimension: incorrect\_information\newline

Consider these points for revising the explanation:\newline
* General: It's important to understand the difference between the different types of energy. Mechanical energy is the energy of motion, while thermal energy is the energy of heat.\newline
* Specific: In the case of the lawnmower, the noise and heat are not a result of the conversion of energy from the fuel to mechanical energy. The noise is a result of the vibration of the engine, while the heat is a result of the friction and combustion of the fuel.\newline

Explanation score: 2\\
    \bottomrule
    \end{tabularx}%
    
    } %small
  \caption{Further examples of explanation critiques from GPT-4 and DS-13B, covering the ``incorrect reasoning'' and ``incorrect information`` dimensions.}
  \label{fig:compare-critiques-gpt4-ds13-p2}%

\end{figure*}%

\begin{figure*}[!htbp]
  {\rule{\textwidth}{0.4pt}}
  \raggedright
  {\small
  \textbf{Question:} Fill in the blank: The tip of James pencil was breaking while writing on the paper sheet. The \_\_\_ is weak. 
(A) pencil 
(B) paper \\
\textbf{Explanation (Llama2-70B):} The word "tip" refers to the pointed end of a pencil. The phrase "breaking while writing on the paper" implies that the pencil was being used to write on the paper. Therefore, the correct answer is (B) paper, as the sentence implies that the paper is weak.\\
\textbf{Answer (Llama2-70B):} (B)  [incorrect, should be (A)]

    \begin{tabularx}{\linewidth}{XX}
    \\
    \toprule
    Critique GPT-4 & Critique DS-13B\\
    \midrule
The explanation states or suggests the following:\newline
 * Main flaw (standalone statement): "Therefore, the correct answer is (B) paper, as the sentence implies that the paper is weak."\newline
 * Dimension: incorrect\_reasoning\newline

Consider these points for revising the explanation:\newline
 * General: Be careful not to misinterpret the cause and effect in a sentence. Just because the pencil tip is breaking while writing on the paper, it doesn't necessarily mean that the paper is weak. It could be that the pencil is weak, which is why its tip is breaking.\newline
 * Specific: In this case, the sentence "The tip of James pencil was breaking while writing on the paper sheet" suggests that the pencil's tip is weak, not the paper. The paper is simply the surface on which the pencil is being used. \newline

Explanation score: 1
&
The explanation states or suggests the following:\newline
* Main flaw (standalone statement): "The correct answer is (B) paper, as the sentence implies that the paper is weak."\newline
* Dimension: misunderstanding\newline

Consider these points for revising the explanation:\newline
* General: Always ensure you understand the question and the context before making a conclusion. Misinterpretation can lead to incorrect answers.\newline
* Specific: The sentence "The tip of James pencil was breaking while writing on the paper" implies that the pencil is weak, not the paper. The pencil is the one that is breaking, which suggests it is weak.\newline

Explanation score: 1\\
    \bottomrule
    \\
    \end{tabularx}%

  \textbf{Question:} Identify best explanation connecting the beginning and ending. Beginning: Karen didn't have a boyfriend for Valentine's Day. Ending: Her brother's thoughtfulness made her Valentine's Day special! 
(A) Karen decided to buy her brother a Valentine's gift. 
(B) Karens little brother handed her a box of chocolates. \\
\textbf{Explanation (GPT-3.5):} 1) The beginning states that Karen didn't have a boyfriend for Valentine's Day.\newline
2) The ending mentions her brother's thoughtfulness made her Valentine's Day special.\newline
3) The connection between the beginning and the ending should involve an action by Karen's brother that made her Valentine's Day special.\newline
4) Giving a box of chocolates is a thoughtful gesture typically associated with Valentine's Day. \\
\textbf{Answer (GPT-3.5):} (B)  [correct]

    \begin{tabularx}{\linewidth}{XX}
    \\
    \toprule
    Critique GPT-4 & Critique DS-13B\\
    \midrule
The explanation states or suggests the following:\newline
 * Main flaw (standalone statement): None\newline
 * Dimension: None\newline

Consider these points for revising the explanation:\newline
 * General: None\newline
 * Specific: None\newline

Explanation score: 5
&
The explanation states or suggests the following:\newline
 * Main flaw (standalone statement): None\newline
 * Dimension: None\newline

Consider these points for revising the explanation:\newline
 * General: None\newline
 * Specific: None\newline

Explanation score: 5\\
    \bottomrule
    \end{tabularx}%
    
    } %small
  \caption{Further examples of explanation critiques from GPT-4 and DS-13B, covering the ``misunderstanding'' and ``None'' dimensions from \modelabb-13B.}
  \label{fig:compare-critiques-gpt4-ds13-p3}%

\end{figure*}%

\begin{table*}[!htbp]
  {\rule{\textwidth}{0.4pt}}
  \raggedright
  {\small
  \textbf{Question:} Which illustrates the correct order of relationships in organisms? 
(A) tissue -> organ system -> organ -> cell 
(B) organ -> cell -> organ system -> tissue 
(C) cell -> tissue -> organ -> organ system 
(D) organ system -> organ -> cell -> tissue\\
\textbf{Explanation (Llama2-70B):} The correct order of relationships in organisms is from the smallest to the largest unit, and it starts with the cell, then tissue, organ, and finally organ system. Therefore, the correct answer is (C) cell -> tissue -> organ -> organ system.\\
\textbf{Answer (Llama2-70B):} (C)  [correct]

    \begin{tabularx}{\linewidth}{XX}
    \\
    \toprule
    Critique DS-13B & Critique DS-7B\\
    \midrule
The explanation states or suggests the following:\newline
 * Main flaw (standalone statement): None\newline
 * Dimension: None\newline

Consider these points for revising the explanation:\newline
 * General: None\newline
 * Specific: None\newline

Explanation score: 5
&
The explanation states or suggests the following:\newline
* Main flaw (standalone statement): "The correct order of relationships in organisms is from the smallest to the largest unit, and it starts with the cell, then tissue, organ, and finally organ system."\newline
* Dimension: incorrect\_information\newline

Consider these points for revising the explanation:\newline
* General: It's important to understand the hierarchy of biological structures and their relationships. The smallest unit is the cell, followed by tissue, organ, and organ system.\newline
* Specific: In the context of this question, the correct order of relationships is from the smallest to the largest unit, which is why option (C) is the correct answer. The order should be: cell, tissue, organ, organ system.\newline

Explanation score: 2\\
    \bottomrule
    \\
    \end{tabularx}%

  \textbf{Question:} A bear cub learns to stay away from unknown bears because 
(A) they are much bigger than the cub 
(B) the other bears look like its mother 
(C) their mother teaches them to keep their distance 
(D) the unknown bears look harmless  \\
\textbf{Explanation (GPT-3.5):} The correct answer is (C) because the sentence states that the bear cub learns to stay away from unknown bears, implying that the mother bear teaches them to keep their distance. Option (A) is incorrect because the size of the unknown bears is not mentioned as the reason for staying away. Option (B) is incorrect because the sentence does not mention the appearance of the unknown bears resembling the cub's mother. Option (D) is incorrect because the sentence does not suggest that the unknown bears look harmless. \\
\textbf{Answer (GPT-3.5):} (D)  [incorrect]

    \begin{tabularx}{\linewidth}{XX}
    \\
    \toprule
    Critique DS-13B & Critique DS-7B\\
    \midrule
The explanation states or suggests the following:\newline
* Main flaw (standalone statement): "The correct answer is (C) because the sentence states that the bear cub learns to stay away from unknown bears, implying that the mother bear teaches them to keep their distance."\newline
* Dimension: inconsistent\_answer\newline

Consider these points for revising the explanation:\newline
* General: Always ensure that your chosen answer aligns with your explanation. If your explanation supports a different answer, it's a sign that you've misunderstood the question or the answer choices.\newline
* Specific: In this case, your explanation correctly identifies that the mother bear teaches the cub to keep its distance, which aligns with option (C). However, you've chosen option (D) as your answer, which doesn't align with your explanation.\newline

Explanation score: 2
&
The explanation states or suggests the following:\newline
* Main flaw (standalone statement): "Option (D) is incorrect because the sentence does not suggest that the unknown bears look harmless."\newline
* Dimension: misunderstanding\newline

Consider these points for revising the explanation:\newline
* General: Always ensure you understand the question and the answer choices correctly. Misunderstanding can lead to incorrect conclusions.\newline
* Specific: In this case, the sentence does suggest that the unknown bears look harmless, which is the correct answer according to the answer sheet. The explanation should focus on why option (C) is the correct answer, not why option (D) is incorrect.\newline

Explanation score: 2\\
   \bottomrule
   \end{tabularx}
    
    } %small
  \caption{Failure examples where DS-7B is producing invalid critiques while DS-13B is producing valid ones.}
  \label{tab:compare-critiques-ds7-ds13}%

\end{table*}%

\eat{
\section{Comprehensive results for DS-13B across dev datasets and student models}
\label{appendix-13b-dataset-model}

%Tables \ref{tab:ds13b_crit_dev_all_datasets_1} and \ref{tab:ds13b_crit_dev_all_datasets_2} show comprehensive results for DS-13B on all 10 datasets of the \dataset{} dev set, across all 4 student models. 
Analyzing comprehensive results for DS-13B on all 10 datasets of the \dataset{} dev set, across all 4 student models, reveals a rich diversity of behavior. E.g., on Physical IQa the models all tend to have more incorrect information than in other RAINBOW datasets, while CosmosQA, especially for GPT-3.5, has a large fraction of incorrect reasoning. Analyzing the individual instances for each dominant explanation flaw will provide more insights into model weaknesses. 
}

\begin{table}[h]
\centering
      \includegraphics[width=\columnwidth]{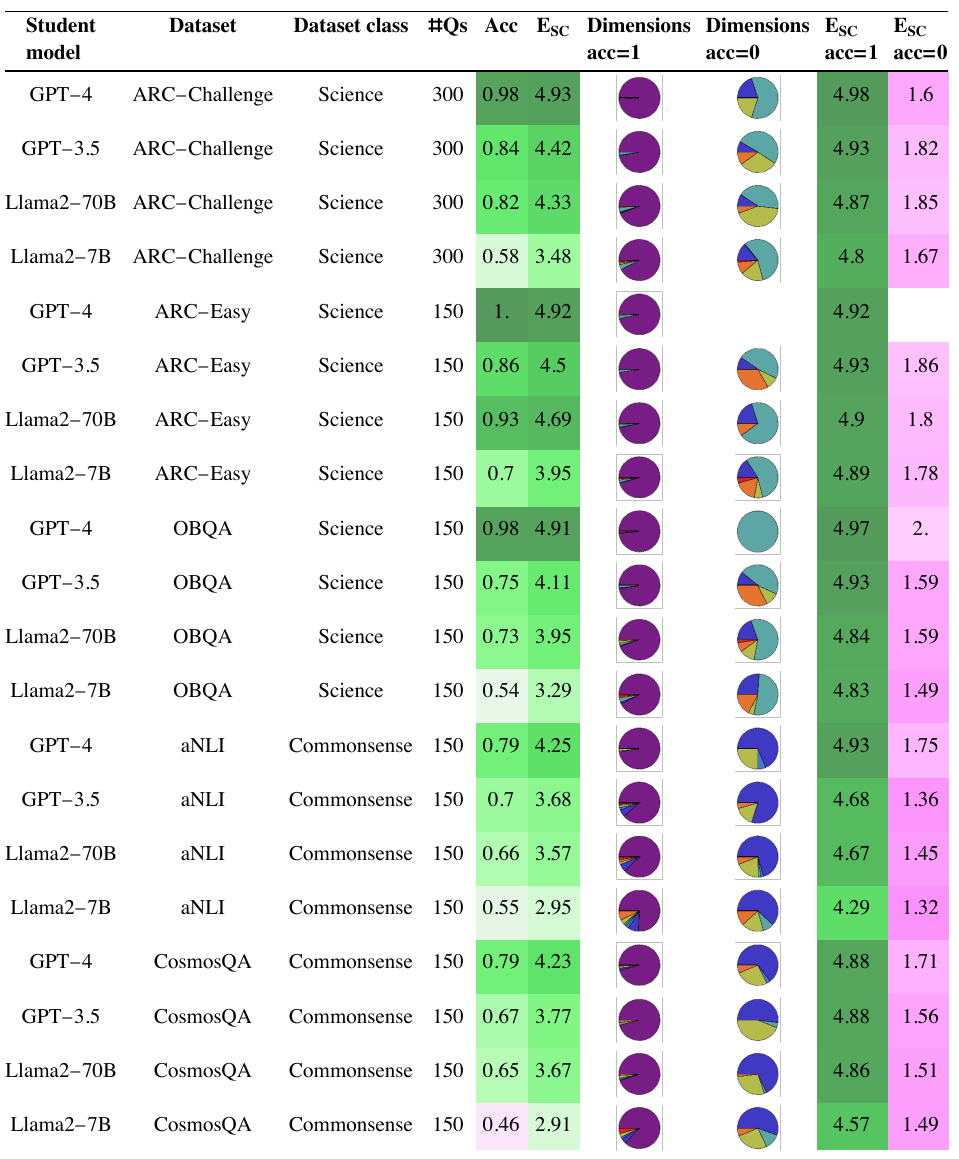}
\caption{Critique analysis from our trained \model{} model DS-13B on the first 5 datasets in the \dataset{} dev set (out of 10 in total). Refer to the legend in Figure \ref{fig:flaw_dimensions_correct_incorrect} for dimensions of flaws. }
\label{tab:ds13b_crit_dev_all_datasets_1}
%\vspace{-4mm}
\end{table}

\begin{table}[h]
\centering
      \includegraphics[width=\columnwidth]{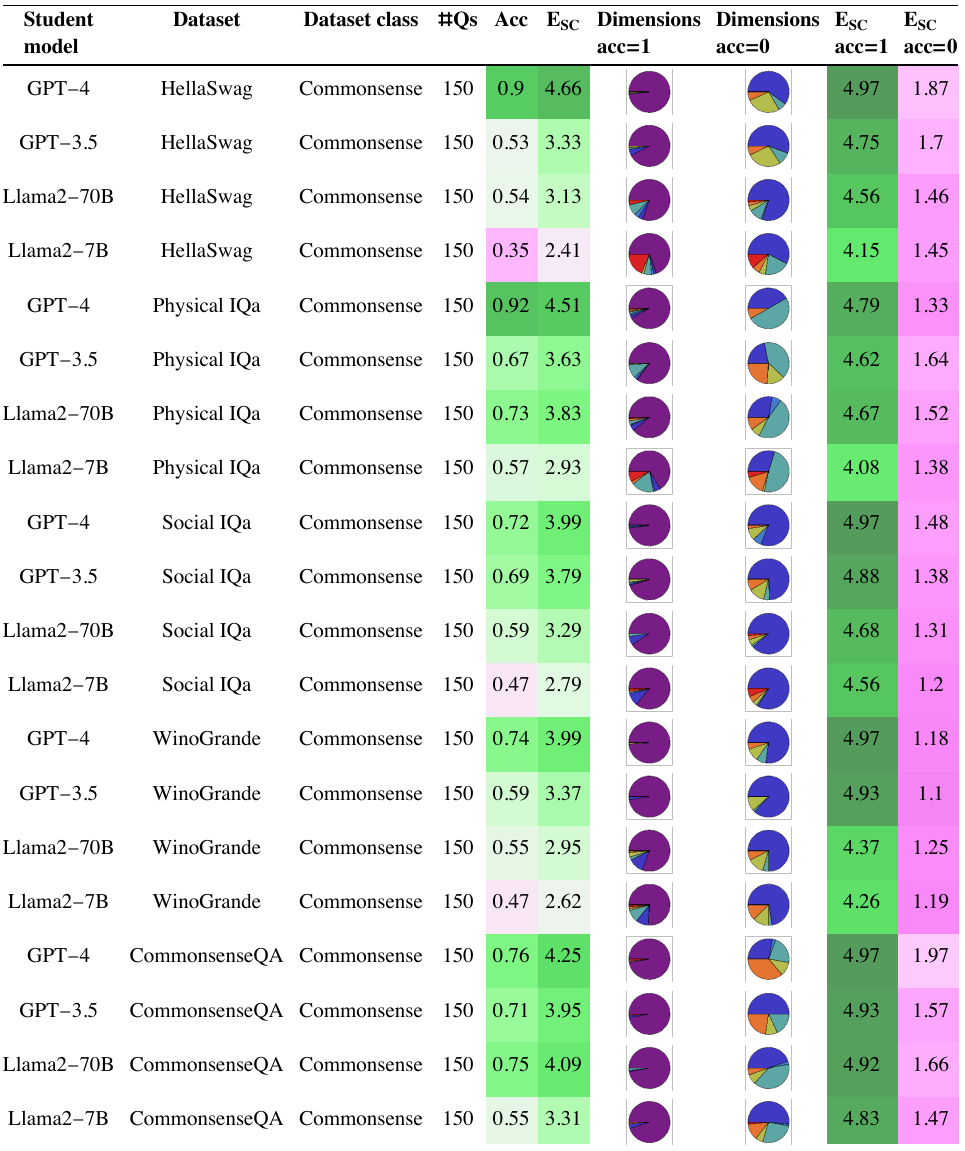}
\caption{Critique analysis from our trained \model{} model DS-13B on the last 5 datasets in the \dataset{} dev set (out of 10 in total). Refer to the legend in Figure \ref{fig:flaw_dimensions_correct_incorrect} for dimensions of flaws. }
\label{tab:ds13b_crit_dev_all_datasets_2}
%\vspace{-4mm}
\end{table}

\section{Prompts for generating different explanations by student models}
\label{appendix-explanation_prompts}

The three different prompts used for generating explanations by the student models can be found in Figures~\ref{fig:zero-shot-prompt}, \ref{fig:few-shot-prompt}, \ref{fig:reasoning-prompt}.

\begin{figure*} [ht]
{\small

{\rule{\textwidth}{0.4pt}}

You are an experienced teacher who can answer the multiple-choice question by explaining why the single selected answer is most correct (and if relevant, why the other answers are incorrect). Produce a concise step-by-step explanation, at most 3 sentences. At the end clearly state the selected answer option, e.g., "Answer: (B)". Use the format:\\
\\
Explanation: ...\\
Answer: ...\\
\\
Here is the question: [[QUESTION]]

{\rule{\textwidth}{0.4pt}}

}
\caption{Zero-shot explanation prompt.}
\label{fig:zero-shot-prompt}
\end{figure*}

\begin{figure*} [ht]
{\small

{\rule{\textwidth}{0.4pt}}
Answer the following multiple-choice questions and explain why each answer choice is either correct or incorrect.\\
\\
Question: Which body system's primary function is the continuation of the species?\\
(A) digestive (B) nervous (C) excretory (D) reproductive\\
Reasoning: The continuation of the species requires children to be born, which is the primary function of the reproductive system (D). The other systems are related to ongoing functioning of the body: The digestive system (A) converts food into useful energy and nutrients. The nervous system (B) controls the communication between the brain and body functions. The excretory system (C) removes excess materials for the body. \\
Answer: (D)\\
\\
Question: Fill in the blank: Amy always raised their voice loudly when they were angry but Christine did not because \_\_\_ had a calmer personality. (A) Amy (B) Christine\\
Reasoning: The blank is referencing the person with the calmer personality, and the sentence states that Amy (A) always raised their voice, while Christine (B) did not, thus Christine is likely the calmer personality.\\
Answer: (B)\\
\\
Question: A scientist cross-pollinates a tomato plant that is resistant to viruses with a tomato plant that produces large tomatoes. This is an example of\\
(A) natural selection. (B) asexual reproduction. (C) selective breeding. (D) genetic engineering.\\
Reasoning: The scientist is selectively breeding (C) plants with desired traits, virus resistance and large tomatoes. Since the scientist interferes, it's not natural selection (A). It is also not asexual reproduction (B) because pollination is used. Because the genes are not directly manipulated, it's not genetic engineering (D). \\
Answer: (C)\\
\\
Question: Which of the following best explains how stems transport water to other parts of the plant?\\
(A) through a chemical called chlorophyll (B) by using photosynthesis (C) through a system of tubes (D) by converting water to food\\
Reasoning: Water is transported in the stem through hollow tubes (C) in the xylem. Chlorophyll (A) and photosynthesis (B) is related to the separate process of converting water to food (D), rather than transportation. \\
Answer: (C)\\
\\
Question: Context: Alex celebrated a birthday with Carson and they both had a fun time. Question: What will happen to Alex? (A) cry hysterically (B) have some food (C) buy a horse\\
Reasoning: The answer food (B) is suggested by the birthday setting, while nothing in the context is suggesting crying hysterically (A) or buying a horse (C).\\
Answer: (B)\\
\\
Question: An engineer can choose from four different materials to make a heat-resistant handle for a stirring spoon. Which material is least likely to get hot when the spoon is placed in boiling water? \\
(A) Ceramic (B) Wood (C) Glass (D) Iron\\
Reasoning: The heat conductivity of the material will affect how hot the spoon gets when placed in boiling water. The heat conductivity of wood (B) is substantially lower than that of ceramic (A), glass (C), and iron (D). \\
Answer: (B)\\
\\
\\
Here is the next question, only answer this question and follow the pattern in the examples, starting with "Reasoning":\\
\\
Question: [[QUESTION]]\\

{\rule{\textwidth}{0.4pt}}
}
\caption{Few-shot explanation prompt.}
\label{fig:few-shot-prompt}
\end{figure*}

\begin{figure*} [ht]
{\small

{\rule{\textwidth}{0.4pt}}
Answer the following multiple-choice question (each with 2 or more choices) by writing down a few verifiable, salient background statements which combine to explain the correct answer. First, here are some examples:\\
\\
Question: Which of the following best explains how stems transport water to other parts of the plant? (A) through a chemical called chlorophyll (B) by using photosynthesis (C) through a system of tubes (D) by converting water to food\\
Reasoning:\\
1) Water is transported in plant stems through hollow tubes in the xylem.\\
Answer: (C) [1]\\
\\
Question: Fill in the blank: Amy always raised their voice loudly when they were angry but Christine did not because \_\_\_ had a calmer personality. (A) Amy (B) Christine\\
Reasoning:\\
1) The sentence states that Amy always raised their voice loudly when they were angry.\\
2) The sentence implies that Christine did not raise their voice loudly when they were angry.\\
3) The person with the calmer personality is the one who did not raise their voice loudly, namely Christine. [1,2]\\
Answer: (B) [3]\\
\\
Question: An engineer can choose from four different materials to make a heat-resistant handle for a stirring spoon. Which material is least likely to get hot when the spoon is placed in boiling water?  (A) Ceramic (B) Wood (C) Glass (D) Iron\\
Reasoning:\\
1) The spoon gets hot when placed in boiling water.\\
2) A handle material with higher heat conductivity will get hotter.\\
3) Wood has substantially lower heat conductivity than ceramic, glass, and iron.\\
Answer: (B) [1,2,3]\\
\\
Question: Which body system's primary function is the continuation of the species? (A) digestive (B) nervous (C) excretory (D) reproductive\\
Reasoning:\\
1) The continuation of the species requires offspring to be produced.\\
2) The primary function of the reproductive system is to produce offspring.\\
Answer: (D) [1,2]\\
\\
Question: Context: Alex celebrated a birthday with Carson and they both had a fun time. Question: What will happen to Alex? (A) cry hysterically (B) have some food (C) buy a horse\\
Reasoning:\\
1) The context mentions a birthday celebration.\\
2) Birthday celebrations typically involve food.\\
3) There is no mention of any event that would cause Alex to cry hysterically or buy a horse.\\
Answer: (B) [1,2,3]\\
\\
Question: A scientist cross-pollinates a tomato plant that is resistant to viruses with a tomato plant that produces large tomatoes. This is an example of (A) natural selection. (B) asexual reproduction. (C) selective breeding. (D) genetic engineering.\\
Reasoning:\\
1) Resistance to viruses is a desired trait.\\
2) Producing large tomatoes is a desired trait.\\
3) The scientist deliberately breeds plants with different, desired traits. [1,2]\\
4) Selective breeding involves choosing parents with particular traits to produce offspring with more desirable traits.\\
Answer: (C) [3,4]\\
\\
\\
Here is the next question, only answer this question and follow the pattern in the examples, starting with "Reasoning":\\
\\
Question: [[QUESTION]]\\

{\rule{\textwidth}{0.4pt}}
}
\caption{Prompt for explanation with reasoning steps.}
\label{fig:reasoning-prompt}
\end{figure*}

\section{Prompt for generating explanations critiques from critique models}
\label{appendix-critique_prompt}

The prompt used for generating critiques by the critique models (GPT-4, DS-13B, and DS-7B) can be found in Figure~\ref{fig:critique-prompt}.

\begin{figure*} [ht]
{\small

{\rule{\textwidth}{0.4pt}}
You are a knowledgeable tutor who gives helpful critique on a given answer and explanation to a question. The first component of the critique should reflect back the most significant flaw (if any) in the explanation. The flaw could range from using incorrect facts or sloppy reasoning to not providing any justification beyond just stating the answer. \\
\\
The flaw should be categorized into one of these dimensions:\\
\\
misunderstanding: apparent misunderstanding of the original question or answer choices\\
lack\_justification: just stating the final answer without any proper or informative justification\\
incorrect\_information: stating incorrect fact(s) or knowledge\\
missing\_information: missing a crucial fact, knowledge, or perspective that should be considered\\
incorrect\_reasoning: an incorrect leap in the reasoning chain\\
incomplete\_reasoning: the reasoning doesn't lead all the way to the answer\\
inconsistent\_answer: the answer doesn't match the conclusion of the explanation\\
irrelevant: using irrelevant or redundant fact(s), knowledge, or reasoning\\
\\
The flaw should be formulated as a standalone erroneous statement, that can be understood without reference to the full question or explanation.\\
\\
When possible, use quote(s) in the reflection statement. In the quote(s), please use "..." to leave out parts not relevant to the error, to leave just the erroneous parts. \\
\\
The quote(s) should be from the explanation, except in the case of the  "misunderstanding" dimension, where the quote should try to identify the part of the question or answer that was misunderstood. \\
\\
If necessary, try to add a brief context to the reflection statement so it can be read standalone without reference to the full question or explanation. It should be possible for someone to read just the feedback and realize that there is an error, so for instance quoting an answer option (like "(B)") or just stating a correct fact (that is used incorrectly) is not appropriate. \\
\\
If an explanation is reasonable, and provides some meaningful evidence supporting the answer, there is no need to be overly picky and possibly misleading, just state "None" as the feedback instead. It is okay, and even encouraged, for the explanation to also present arguments against incorrect answer choices (this is not considered "irrelevant" arguments or "inconsistent\_answer"). Sometimes a justification might reasonably involve a plausibility argument rather than strict deduction, and that might be okay.\\
\\
\\
The second component of the critique should provide statements toward fixing this flaw in the explanation. First, try to provide a general statement that addresses a likely misconception or confusion behind the flaw. Formulate this so it can be reused independently of this particular question. Next, provide a more specific statement to consider when revising the explanation.\\
\\
Finally, please grade the explanation on a scale from 0 to 5, with the scoring guidelines:\\
\\
0: Very wrong explanation\\
1: Has something useful in the explanation that's relevant to the question\\
2: Makes some valid points, but is mostly incorrect\\
3: Has many correct elements, but with significant flaws\\
4: Mostly correct, with a minor flaw\\
5: Completely correct\\
\\
Please use the following format for the critique (use "None" if there is no feedback for a particular line):\\
\\
The explanation states or suggests the following:\\
 * Main flaw (standalone statement): <flaw>\\
 * Dimension: <dimension>\\
\\
Consider these points for revising the explanation:\\
 * General: <reusable statement>\\
 * Specific: <question-specific statement> \\
\\
Explanation score: <score>\\
\\
Here is the question with the given answer and explanation:\\
\\
Question: [[QUESTION]]\\
Correct answer (according to answer sheet): ([[ANSWERKEY]])\\
Given answer: ([[PREDICTEDANSWER]])\\
Given explanation: [[EXPLANATION]]

{\rule{\textwidth}{0.4pt}}

}
\caption{Prompt for generating critiques.}
\label{fig:critique-prompt}
\end{figure*}

\section{Training hyperparameters}
\label{appendix-training-hyperparams}

The fine-tuning of the \model{} DS-7B and DS-13B critique models were done on NVIDIA RTX A6000 GPUs, using the open-instruct code base.\footnote{https://github.com/allenai/open-instruct} The base models were Llama2-7B-Chat and Llama2-13B-Chat respectively. Low-Rank Adaptation (LoRA) \citep{hu2021lora} was used for parameter-efficient fine-tuning. The hyperparameters included: batch size = 32, lora rank = 64, lora alpha = 16, lora dropout = 0.1, learning rate = 1e-4, lr schedule = linear, warmup ratio = 0.03, weight decay = 0, flash attention. For the 13B model we also used QLoRA \cite{dettmers2023qlora}. A total of around 100 GPU hours was used between fine-tuning and inference for generating the \dataset{} dev set critiques.

\end{document}